\documentclass[journal]{IEEEtran}
\IEEEoverridecommandlockouts
\usepackage{graphicx}
\def\BibTeX{{\rm B\kern-.05em{\sc i\kern-.025em b}\kern-.08em
    T\kern-.1667em\lower.7ex\hbox{E}\kern-.125emX}}
\usepackage[font=small,labelfont=bf,tableposition=top]{caption}

\usepackage{blindtext}
\usepackage{caption}

\let\oldtwocolumn\twocolumn
\renewcommand\twocolumn[1][]{%
    \oldtwocolumn[{#1}{
    \begin{center}
        \centering
        \includegraphics[width=0.99\textwidth]{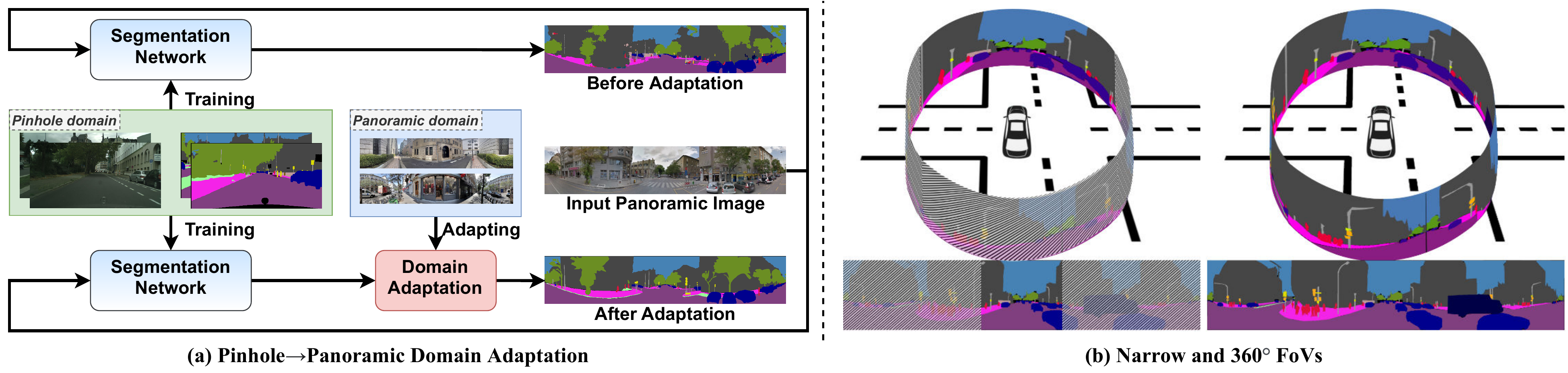}
        \captionof{figure}{(a) An overview of the formalized task of domain adaptation for panoramic semantic segmentation. The \emph{source} domain (green background) contains \emph{pinhole} images with semantic annotations, while the \emph{target} domain (blue background) contains \emph{panoramic} images without annotations. (b) FoV comparison between pinhole forward-view and $360^\circ$ panoramic surround-view imaging of self-driving scenes.}
        \label{fig.introduce}
    \end{center}
    }]
}

\usepackage{amsmath,amssymb,amsfonts}
\usepackage[inline]{enumitem}
\usepackage{algorithmic}
\usepackage{graphicx}
\usepackage{textcomp}

\usepackage{booktabs}
\usepackage{multirow}
\usepackage{hyperref}
\hypersetup{
colorlinks=true,linkcolor=red,citecolor=blue,urlcolor=black}

\usepackage[utf8]{inputenc}
\usepackage[T1]{fontenc}
\usepackage{float}
\usepackage{adjustbox}
\usepackage{array}
\usepackage{etoolbox}
\usepackage{dsfont}
\usepackage{xcolor}
\definecolor{revised_color}{HTML}{0066CC}

\usepackage{subcaption}

\usepackage[capitalise]{cleveref}

\begin{document}

\title{Transfer beyond the Field of View: Dense Panoramic Semantic Segmentation via Unsupervised Domain Adaptation}
\author{Jiaming Zhang$^{1}$, Chaoxiang Ma$^{1}$, Kailun Yang$^{1}$, Alina Roitberg$^{1}$, Kunyu Peng$^{1}$, and Rainer Stiefelhagen$^{1}$
\thanks{This work was supported in part through the AccessibleMaps project by the Federal Ministry of Labor and Social Affairs (BMAS) under the Grant No. 01KM151112, in part by the University of Excellence through the ``KIT Future Fields'' project, and in part by Hangzhou SurImage Company Ltd.
\emph{(Corresponding author: Kailun Yang.)}}
\thanks{$^{1}$Authors are with Institute for Anthropomatics and Robotics, Karlsruhe Institute of Technology, Germany (e-mail: jiaming.zhang@kit.edu, chaoxiang.ma.1024@gmail.com, kailun.yang@kit.edu, alina.roitberg@kit.edu, kunyu.peng@kit.edu, rainer.stiefelhagen@kit.edu).}
\thanks{Code and dataset will be made publicly available at: https://github.com/chma1024/DensePASS}
}

\maketitle

\begin{abstract}
Autonomous vehicles clearly benefit from the expanded Field of View (FoV) of $360^\circ$ sensors, but modern semantic segmentation approaches rely heavily on annotated training data which is rarely available for \emph{panoramic} images. We look at this problem from the perspective of domain adaptation and bring \emph{panoramic} semantic segmentation to a setting, where labelled training data originates from a different distribution of conventional \emph{pinhole} camera images. To achieve this, we formalize the task of unsupervised domain adaptation for panoramic semantic segmentation and collect \textsc{DensePASS} - a novel densely annotated dataset for panoramic segmentation under cross-domain conditions, specifically built to study the \textsc{Pinhole$\rightarrow$Panoramic} domain shift and accompanied with pinhole camera training examples obtained from Cityscapes. \textsc{DensePASS} covers both, labelled- and unlabelled $360^\circ$ images, with the labelled data comprising $19$ classes which explicitly fit the categories available in the source (\textit{i.e.} pinhole) domain. Since data-driven models are especially susceptible to changes in data distribution, we introduce P2PDA - a generic framework for \textsc{Pinhole$\rightarrow$Panoramic} semantic segmentation which addresses the challenge of domain divergence with different variants of attention-augmented domain adaptation modules, enabling the transfer in output-, feature-, and feature confidence spaces. P2PDA intertwines uncertainty-aware adaptation using confidence values regulated on-the-fly through attention heads with discrepant predictions. Our framework facilitates context exchange when learning domain correspondences and dramatically improves the adaptation performance of accuracy- and efficiency-focused models. Comprehensive experiments verify that our framework clearly surpasses unsupervised domain adaptation- and specialized panoramic segmentation approaches as well as state-of-the-art semantic segmentation methods.

\end{abstract}


\IEEEpeerreviewmaketitle

\section{Introduction}
\IEEEPARstart{S}{emantic} segmentation assigns a category label to every image pixel~\cite{fcn} and is vital for perception of autonomous vehicles as it enables to locate key entities of a driving scene, such as \emph{road}, \emph{sidewalk} or \emph{person}~\cite{peng2021mass}. 
Although the semantic segmentation accuracy has increased at a rapid pace thanks to the resilience of Convolutional Neural Networks~(CNNs),
most of the previously published frameworks are developed under the assumption that the driving scene images are captured with a \emph{pinhole} camera~\cite{cityscapes}.
However, the comparably narrow Field of View~(FoV) largely limits the perception capacity.
Mounting multiple sensors can mitigate this issue, but requires additional data fusion mechanisms in return~\cite{restricted,camera_lidar}.
Recently, leveraging a single \emph{panoramic} camera, which offers a unified $360^\circ$ perception of the driving environment, as depicted in Fig.~\ref{fig.introduce}(b), started to gain attention as a novel alternative way for expanding the FoV~\cite{can_we_pass,pass}.

Unfortunately, the scarcity of pixel-wise annotation of panoramic images still hinders the progress of the semantic segmentation research for such data.
At the same time, recent progress in the area of Domain Adaptation (DA) has lead to highly effective techniques, giving a new perspective to complement the insufficient coverage of training data in driving scenarios, \textit{e.g.}, at the nighttime~\cite{bridging} or in accident scenes~\cite{issafe}.
In this paper, we look at the problem of label-scarce panoramic segmentation through the lens of domain adaptation and target knowledge transfer from the significantly larger datasets available in the domain of standard images. 
To this intent, we formalize the task of unsupervised domain adaptation for panoramic segmentation in a novel Dense PAnoramic Semantic Segmentation (\textsc{DensePASS}) benchmark, where images in the label-scarce panoramic \emph{target} domain are handled by adapting from data of the label-rich pinhole \emph{source} domain (Fig.~\ref{fig.introduce}(a) provides an overview of the formalized task). 

To foster research of \emph{panoramic semantic segmentation under cross-domain conditions}, we introduce \textsc{DensePASS} -- a new dataset covering $360^\circ$ images captured from all around the globe to facilitate diversity.
To provide credible quantitative evaluation, our benchmark comprises (1) an unlabelled panoramic training set used for optimization of the domain adaptation model and (2) a panoramic test set manually labelled with $19$ classes defined in accordance to Cityscapes~\cite{cityscapes}, a dataset with pinhole images which we use as the label-rich training data from the source domain. 

Unfortunately, a straightforward transfer of models trained on pinhole images to panoramic data often results in a significant drop in accuracy, as the panoramic layout of images passed through the equirectangular projection deviates from the standard pinhole camera data.
For example, as shown in Fig.~\ref{fig.introduce}(b), panoramic images have longer horizontal distribution or geometric distortion on both sides of the viewing direction, resulting in a considerable domain shift.

To meet the challenge of label-scarce panoramic segmentation by learning from label-rich pinhole image datasets, we implement P2PDA - a generic framework for the \emph{Pinhole to Panoramic Domain Adaptation}.
We examine different domain adaptation mechanisms: (i) \emph{Segmentation Domain Adaptation Module~(SDAM)}, (ii) \emph{Attentional Domain Adaptation Module~(ADAM)}, (iii) \emph{Regional Context Domain Adaptation Module~(RCDAM)}, and (iv) \emph{Feature Confidence Domain Adaptation Module~(FCDAM)}.

Specifically, the proposed \emph{SDAM} module allows greater flexibility than the previous DA methods~\cite{adaptsegnet,all_about_structure} used in the output-space as it can be plugged in at different feature levels.
Another challenge is  learning robust representations, which are not only discriminative for various categories with similar appearances, but also connect regions of the same category at diverse locations across the $360^\circ$.
To address this, we leverage the progress of attention-based models~\cite{nonlocal,danet,fanet} and propose the \emph{ADAM} module for capturing long-range dependencies and positional relations.
Besides, the \emph{RCDAM} module addresses the horizontal distribution of panoramic images and obtains region-level context in order to effectively resist the geometric distortion caused by the equirectangular projection. Lastly, the \emph{FCDAM} module enforces the backbone model to generate features with higher confidence in the respective domain.
These modules can be flexibly activated through our P2PDA framework individually or jointly.

According to our observations, despite the different FoVs, the predictions of the two domains in the output-space still have similar contextual distributions. For instance, \emph{sky} is still distributed in the upper part of the image, \emph{vegetation} in the middle part, the \emph{road}-related mostly in the lower part, etc. Therefore, we advocate \emph{multi-level alignment} by deploying our building blocks for domain adaptation in different spaces. 
Through the  complementary nature of the DA modules and their combination through P2PDA, we achieve simultaneous adaptation in the output-, feature-, and feature confidence spaces.
Furthermore, P2PDA intertwines an uncertainty-aware adaptation phase using confident online panoramic pseudo-labels, where the uncertainty estimation is regulated by attention heads with discrepant predictions.

Extensive evaluation of the \textsc{Pinhole$\rightarrow$Panoramic} transfer demonstrates the effectiveness of our framework, significantly boosting the domain adaptation performance of both accuracy- and efficiency-oriented models~\cite{danet,fanet,erfnet}.
Finally, the proposed P2PDA strategy outperforms recent panoramic segmentation and domain adaptation frameworks, with our P2PDA-driven DANet surpassing $>20$ state-of-the-art CNN- and transformer-based  segmentation models.

This work is the extension of our conference paper~\cite{densepass_itsc}, extended with several domain adaptation modules, a detailed description of the DensePASS dataset, and an extended set of experiments and analysis. In summary, our main contributions are summarized as follows:
\begin{itemize}
\item We create and publicly release \textsc{DensePASS} -- a new benchmark for panoramic semantic segmentation collected from locations all around the world and densely annotated with $19$ classes in accordance to the pinhole camera dataset Cityscapes to enable proper \textsc{Pinhole$\rightarrow$Panoramic} evaluation.
\item We propose a generic P2PDA framework and investigate various DA modules both in a separate and joint manner, validating their effectiveness with various networks designed for self-driving scene segmentation.
\item We advocate attentional domain adaptation by integrating attention-augmented adversarial- and attention-regulated self-learning adaptation, verifying that uncertainty-aware distillation with adapted knowledge can further boost the DA performance significantly. 
\item With the DANet~\cite{danet} baseline, our P2PDA framework achieves $+13.5\%$ and $+16.2\%$ mIoU gains by adapting from Cityscapes~\cite{cityscapes} and further adding WildDash~\cite{wilddash}.
\end{itemize}

\section{Related Work}

\subsection{Semantic Segmentation and Self-attention Modules}

Semantic segmentation has progressed almost exponentially due to the architectural advances of Fully Convolutional Networks~\cite{fcn} and the increasing amount of available real-life training data~\cite{deep_multimodal}.
Models such as PSPNet~\cite{pspnet} and DeepLabV3+~\cite{deeplabv3+} attain great accuracy improvements on conventional benchmarks  by harvesting multi-scale feature representations using atrous convolution or pyramid pooling.
In~\cite{panopticfpn}, a dense prediction branch, Semantic-FPN, is attached on top of the feature pyramid, individually for semantic segmentation.
While these composite architectures achieve fine-grained precise segmentation, efficient and compact networks like ERFNet~\cite{erfnet}, SwiftNet~\cite{swiftnet}, and Fast-SCNN~\cite{fastscnn} aim to perform both fast and accurate segmentation.
For self-driving scene parsing, most of existing benchmark datasets are captured by pinhole cameras, for example Cityscapes~\cite{cityscapes}, Mapillary Vistas~\cite{mapillary}, ApolloScape~\cite{apolloscape}, BDD~\cite{bdd}, IDD~\cite{idd}, KITTI~\cite{kitti360}, and WildDash~\cite{wilddash}.
Despite large receptive fields, most segmentation algorithms are driven by data availability and are therefore designed for standard narrow-FoV images.

Another line of work leverages \emph{self-attention modules}~\cite{attention}, which automatically weighs input positions (\textit{i.e.,} temporal~\cite{attention} or spatial~\cite{nonlocal}), gains increasing attention in the field. Such mechanisms are broadly used for capturing long-range contextual dependencies, which are crucial for dense-pixel prediction tasks~\cite{zhang2021trans4trans}.
The success of attention mechanisms in visual recognition~\cite{nonlocal}, leads to various explorations in semantic segmentation works focused on both, accuracy-oriented networks~\cite{danet,ranet,ocrnet} and efficiency-oriented networks~\cite{fanet,omnirange,attanet}.
For instance, Fast Attention Network (FANet)~\cite{fanet} uses a fast attention mechanism by replacing the Softmax normalization with cosine similarity, whereas Dual Attention Network (DANet)~\cite{danet} appends position- and channel attention modules to model semantic associations between any two pixels or channels.
Other prominent techniques include criss-cross attention~\cite{ccnet} and disentangled non-local attention blocks~\cite{omnirange,dnl,annn}.
In particular, ECANet~\cite{omnirange} concurrently highlights horizontally-driven dependencies and collects omni-range contextual information for large-FoV semantic segmentation.

We leverage such attention principles to mitigate the domain shift by highlighting regional context and present a cross-domain segmentation framework with \emph{attentional domain adaptation modules}.
We experiment with both, accuracy- and efficiency-focused networks~\cite{danet,fanet,erfnet} as the segmentation architecture, and demonstrate the consistent effectiveness of our adaptation modules for bringing standard semantic segmentation models to panoramic imagery.

\subsection{Semantic Segmentation for $360^\circ$ Panoramic Images}

Segmentation of panoramic data, which is often captured through distortion-pronounced fisheye lenses~\cite{fisheye,universal,adaptable_deformable} or multiple surround-view cameras~\cite{crossview,omnidet,hu2021fiery}, is challenging as it entails a set of hard tasks like distortion elimination, camera synchronization and calibration, as well as data fusion, resulting in higher latency and complexity.
Yang~\textit{et al.} introduce the PASS~\cite{pass}
and the DS-PASS~\cite{dspass} frameworks which naturally mitigate the effect of distortions by using a single-shot panoramic annular lens system, but come with an expensive memory- and computation cost, as it requires separating the panorama into multiple partitions for predictions, each resembling a narrow-FoV pinhole image.
This is significantly improved by the OOSS framework~\cite{ooss} through multi-source omni-supervised learning.
The latest advancements include frameworks focusing on dimension-wise positional priors~\cite{wildpass}, omni-range contextual dependencies~\cite{omnirange}, or leveraging contrastive pixel-propagation pre-training~\cite{pps}.

All previous frameworks~\cite{pass,omnirange,pps} are developed under the assumption that the labelled training data are implicitly or partially available in the target domain of panoramic segmentation.
Since panoramic datasets are comparably small in size, we argue, that \emph{panoramic} segmentation might strongly benefit from the significantly larger datasets available in the domain of \emph{pinhole} camera image segmentation.
While synthetic omnidirectional datasets exist~\cite{orientation,synthetic,omniscape,cortinhal2021semantics,orhan2021semantic}, they lack diversity and realism present in the large-scale pinhole image collections with rich ontologies~\cite{cityscapes,wilddash}.
In this work, we look at panoramic segmentation from a \emph{domain adaptation perspective} and introduce the \textsc{DensePASS} dataset covering images with $19$ annotated categories available in both, standard- and panoramic domains.
We introduce a framework for unsupervised domain adaptation for panoramic semantic segmentation, where we combine prominent segmentation approaches with attention-based domain adaptation modules involving attention-augmented adversarial learning and attention-regulated self-training.

\subsection{Unsupervised Domain Adaptive Semantic Segmentation}
To tackle model generalization to new scenes, \emph{domain adaptation} became an increasingly popular topic, offering new ways to complement the insufficient coverage of training data in driving scenarios, \textit{e.g.}, at the nighttime~\cite{bridging,see_clearer_at_night,nighttime_road_scene_parsing,rainy_night},
or in accident scenes~\cite{issafe}.
As neural networks are especially vulnerable to changes in data distribution, various domain adaptation frameworks have been proposed to  address this challenge with the  most common practices being self-training~\cite{curriculum_da,pycda,crst,rectifying,proda} and adversarial learning~\cite{adaptsegnet,all_about_structure,fcnsinthewild,cycada,clan}. 
Self-training based  methods are grounded in iterative improvement, \textit{e.g.}, through pseudo-labels.
PyCDA~\cite{pycda} views self-training from the perspective of curriculum adaptation by introducing a pyramid curriculum consisting of various properties about the target domain. 
CRST~\cite{crst} is formulated as a regularized self-training framework, which takes the confident predictions in target domains as soft pseudo-labels for segmentation network retraining.
Zheng and Yang~\cite{rectifying} leverage uncertainty estimation to enable automatic thresholding of noisy pseudo-labels for unsupervised segmentation adaptation.
The recent ProDA~\cite{proda} resorts to representative prototypes, namely class-wise cluster centroids, to gradually rectify soft pseudo-labels and produce a compact target feature structure.

The second prominent group of approaches is driven by Generative Adversarial Networks (GANs)~\cite{gan}, which learn domain translations, \textit{e.g.}, via image-to-image conversions~\cite{bridging,see_clearer_at_night,cycada}, feature alignment~\cite{fcnsinthewild}, or semantic layout matching~\cite{adaptsegnet,all_about_structure,content_consistent_matching_da,contextual_relation_consistent_da}.
To utilize the significant amount of \textit{source}-\textit{target} similarities in the resulting segmentation masks, Tsai~\textit{et al.}~\cite{adaptsegnet} align the domains in output-space via adversarial learning (AdaptSegNet).
Chang~\textit{et al.}~\cite{all_about_structure} construct the DISE framework which extracts domain-invariant structure and domain-specific texture information to reduce source-target discrepancies.
More recent works further prioritize category-level alignment (CLAN)~\cite{clan}, minimize adversarial entropy~\cite{advent}, or perform affinity-space domain adaptation~\cite{affinity_space_da}.

\begin{figure*}[!h]
  \centering
  \includegraphics[width= \textwidth]{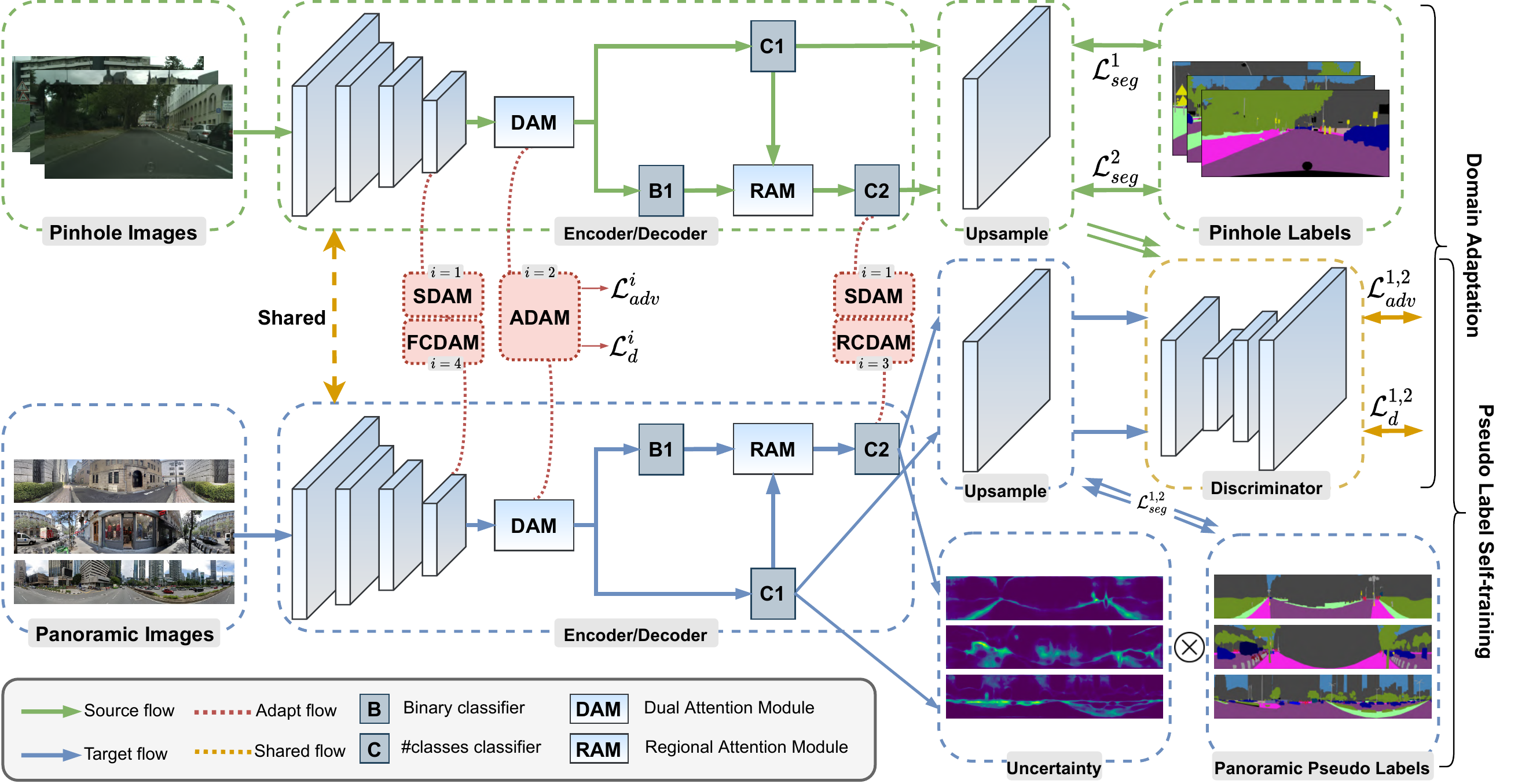}
  \vskip-1ex
  \caption{Diagram of the P2PDA framework. The shared backbone is an encoder-decoder segmentation network (\textit{i.e.} DANet). The SDAM module is applied on the high-level feature-space or output-space and the FCDAM on the feature confidence space, while ADAM is performed on the dual attended feature-space and RCDAM on the output-space after the region attention module. Each of these four modules includes an individual discriminator and an adversarial loss. In the second stage of pseudo-label self-supervised learning, the uncertainty map is calculated based on the C1 and C2 predictions and used for element-wise multiplication with pseudo-labels as an online selection of pseudo-labels.
  }
  \vskip-3ex
  \label{fig.overview}
\end{figure*}

Domain adaptation also increasingly leverages attention mechanisms by, \textit{e.g.}, attending feature maps to highlight transferable areas~\cite{tada,safe},
minimizing the source-target distribution divergence of the attention maps~\cite{source_free}, or using multiple cross-domain attention modules for obtaining context dependencies from both local and global perspectives~\cite{contextaware}.
mDALU~\cite{gong2020mdalu} performs partially-supervised learning and attention-guided fusion to tackle a specific problem of multi-source domain adaptation and label unification.
In this work, we specifically focus on domain transfer for \emph{panoramic} semantic segmentation, which differs from the standard \emph{pinhole} images in several important aspects, such as discontinuous boundaries and distorted objects.
To exploit long-range correlations between pixels and semantic regions, we extend AdaptSegNet~\cite{adaptsegnet} with attention-augmented modules in multiple levels and a regional context exchange, simultaneously adapting in the output-, feature-, and feature confidence spaces.
This enables direct information exchange across the whole FoV, mitigating the influence of discrepancy in positional priors and local distortions, which is vital for the \textsc{Pinhole$\rightarrow$Panoramic} transfer.
Moreover, the proposed framework intertwines an uncertainty-aware adaptation phase using confident online panoramic labels regulated by attention modules, which further reduces the domain gap.

\section{P2PDA: Proposed Framework}

In this work, we introduce a generic framework for $360^\circ$ perception of self-driving scenes by learning to adapt semantic segmentation networks from a label-rich source domain of standard pinhole camera images to the unlabelled target domain of panoramic data.
Conceptually, our framework covers an encoder-decoder based semantic segmentation network and four different building blocks for domain alignment: Segmentation domain adaptation module (SDAM), Attentional domain adaptation module (ADAM), Regional context domain adaptation module (RCDAM), and Feature confidence domain adaptation module (FCDAM), which we place at two different network stages: after and before the decoder of the segmentation network.
In the following, we  give an overview of the proposed framework (Sec. \ref{sec:overview}) and present the integrated domain adaptation modules in detail (Sec. \ref{sec:da-modules}).

\subsection{Framework Overview}
\label{sec:overview}

\textbf{Attention-augmented adversarial adaptation.}
Our framework builds on the AdaptSegNet architecture~\cite{adaptsegnet} extending it with multiple variants of region- or attention-augmented DA modules plugged in at different network depths with an overview provided in Fig.~\ref{fig.overview}. 
The main components of our framework are a weight-shared segmentation network \textbf{G} with attention modules and multiple DA modules equipped with the corresponding discriminators \textbf{D}.
For unsupervised domain adaptation methods, only the source domain dataset $\mathcal{D}_s=\{(x_s, y_s)|x_s \in \mathbb{R}^{H_s\times W_s\times 3}, y_s \in \mathbb{R}^{H_s\times W_s\times 1}\}$ and the unlabelled target domain dataset $\mathcal{D}_t=\{(x_t)|x_t \in \mathbb{R}^{H_t\times W_t\times 3}\}$ are given, where $x_s$ and $x_t$ denote the input images from source and target domains, and $y_s$ are the ground truth labels in source domain. We note that ($H_s$, $W_s$) and ($H_t$, $W_t$) are the height and width of the source and target images, respectively.

For ease of understanding, we only list the formulas for a single classifier on \textbf{G} and single \textbf{D}. For multiple \textbf{G} or \textbf{D}, they will be combined through specific hyper-parameters. First, the \emph{source} domain images $x_s$ are fed into the segmentation network \textbf{G} (also referred to as the generator) to create prediction results $\tilde{y}_s=G(x_s)$ and the source ground-truth labels $y_s$ are used to compute the segmentation loss $\mathcal{L}_{seg}$ as follows:
\begin{equation}\label{eqn:seg_loss}
    \mathcal{L}_{seg}(G) = \mathbb{E}\left[ \ell (G(x_s), y_s)\right],
\end{equation}
where $E[\cdot]$ is the statistical expectation and $\ell(\cdot, \cdot)$ is the standard cross entropy loss. 

Next, the discriminator \textbf{D} is trained with the binary objective to distinguish between the source and target domains of the input, so that the discriminator loss is formulated as:
\begin{equation}
    \mathcal{L}_{d}(D) = \mathbb{E}\left[ \ell (D(G(x_s)), 0)\right] + \mathbb{E}\left[ \ell (D(G(x_t)), 1)\right],
\end{equation}\label{eqn:d_loss}
where the $\ell(\cdot, \cdot)$ is binary cross entropy, with $0$ and $1$ being the two-class labels (panoramic and pinhole).

Then, to enforce the generator \textbf{G} to align the distribution of $\tilde{y}_t$ closer to $\tilde{y}_s$, the prediction results $\tilde{y}_t=G(x_t)$ for the target domain is directly used to estimate the adversarial loss, which is updated alongside with $\mathcal{L}_{seg}$ and is formulated as:
\begin{equation}\label{eqn:adv_loss}
    \mathcal{L}_{adv}(G) = \mathbb{E}\left[ \ell (D(G(x_t)), 0)\right].
\end{equation} 
The adversarial loss is high if the discriminator prediction is correct, 
so the adversarial loss facilitates generation of segmentation masks in the target domain which successfully ``fool'' the discriminator.
In other words, the discriminators are trained to distinguish between the source and target domains with $\mathcal{L}_{d}(D)$, while the segmentation network $\textbf{G}$ is trained to 1) correctly segment the images from the source domain with $\mathcal{L}_{seg}$, and 2) ``fool'' the discriminator by making the target domain data indistinguishable from the source domain data.

The join loss from Eqn.~\eqref{eqn:seg_loss} and Eqn.~\eqref{eqn:adv_loss} used to train the generator \textbf{G} becomes:
\begin{equation}\label{eqn:G_loss}
 \mathcal{L}(G) = \lambda_{seg}\mathcal{L}_{seg}(G) + \lambda_{adv}\mathcal{L}_{adv}(G),
\end{equation}
where $\lambda_{adv}$ and $\lambda_{seg}$ are weights used to balance the domain adaptation and semantic segmentation losses. To perform end-to-end training for multiple classifiers of \textbf{G} and multiple \textbf{D}, our final loss function is denoted as:
\begin{equation}\label{eqn:finale_loss}
\begin{aligned}
    \mathcal{L}(G, D) &= \sum_{i}\lambda_{seg}^i\mathcal{L}_{seg}^i(G) + \sum_{i}\lambda_{adv}^i\mathcal{L}_{adv}^i(G) \\&+ \sum_{i}\lambda_{d}^i\mathcal{L}_{d}^i(D).
\end{aligned}
\end{equation}

\textbf{Attention-regulated self-learning adaptation.}
Our network can readily generate highly qualified segmentation masks on panoramic images, after the main stage of multi-level alignment through the domain adaptation modules described in Sec. \ref{sec:da-modules}.
Our next goal is to advance the training procedure by leveraging the inherent knowledge present in the pixel-wise predictions obtained after the first training stage as \emph{panoramic pseudo-labels}.
To achieve this, P2PDA intertwines an uncertainty-aware domain adaptation stage by using panoramic pseudo-labels with high confidences on-the-fly, therefore improving the prediction in iterative fashion.
In this stage, the source images are replaced by the self-supervised panoramic images, \textit{i.e.}, the predictions are used to refine the model itself.
The key idea of this training stage is to employ multiple classifiers with attention heads naturally encouraged to produce \emph{discrepant} predictions in order to assess the uncertainty of the pseudo-labels.
First, we estimate the uncertainty map by using  the variance operation on predictions produced with two different classifiers with disparate attention modules as in~\cite{danet,ranet}. 
Then, we apply element-wise multiplication of the pseudo-labels with the resulting uncertainty map and, finally, we threshold the resulting value to obtain the \emph{certain} pseudo-labels.
An overview of our uncertainty-driven self-training is illustrated in the bottom part of Fig.~\ref{fig.overview}.

In the next section, we describe our building blocks for domain alignment and adaptation.

\subsection{Domain Adaptation Modules}
\label{sec:da-modules}

\begin{figure}[!t]
  \centering
  \includegraphics[width=0.485\textwidth]{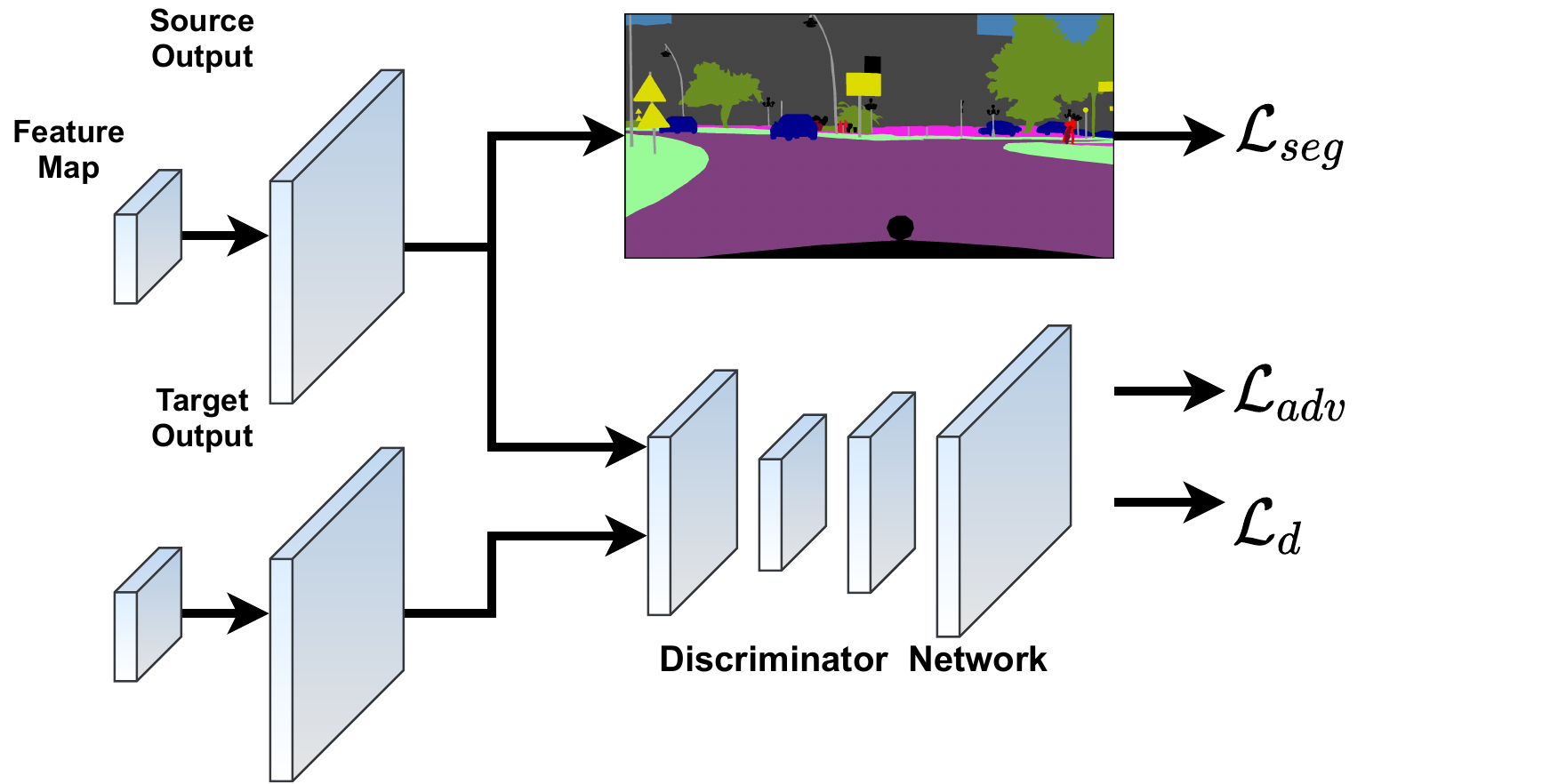}
  \caption{An overview of the SDAM module.}
  \label{fig.overview.segmentation.da}
  \vskip-3ex
\end{figure}

\textbf{Segmentation domain adaptation module.}
Our initial domain adaptation module SDAM is derived from AdaptSegNet and attempts to match the source and target segmentation output (the module is illustrated in Fig.~\ref{fig.overview.segmentation.da}).
After a segmentation network forward pass with images from  both domains ($x_s$ and $x_t$), feature maps of the both representations are used as input to the discriminator \textbf{D} which learns to distinguish the domain with $\mathcal{L}_{d}(D)$, while the segmentation network $G$ learns to  segment the pinhole images with $\mathcal{L}_{seg}(G)$ and align the domains with $\mathcal{L}_{adv}(G)$.
SDAM learns a \textsc{Pinhole$\rightarrow$Panoramic} domain adaptation model at \emph{multiple levels jointly} within our P2PDA framework, as shown in Fig.~\ref{fig.overview}.

\begin{figure}[!t]
  \centering
  \includegraphics[width=0.45\textwidth]{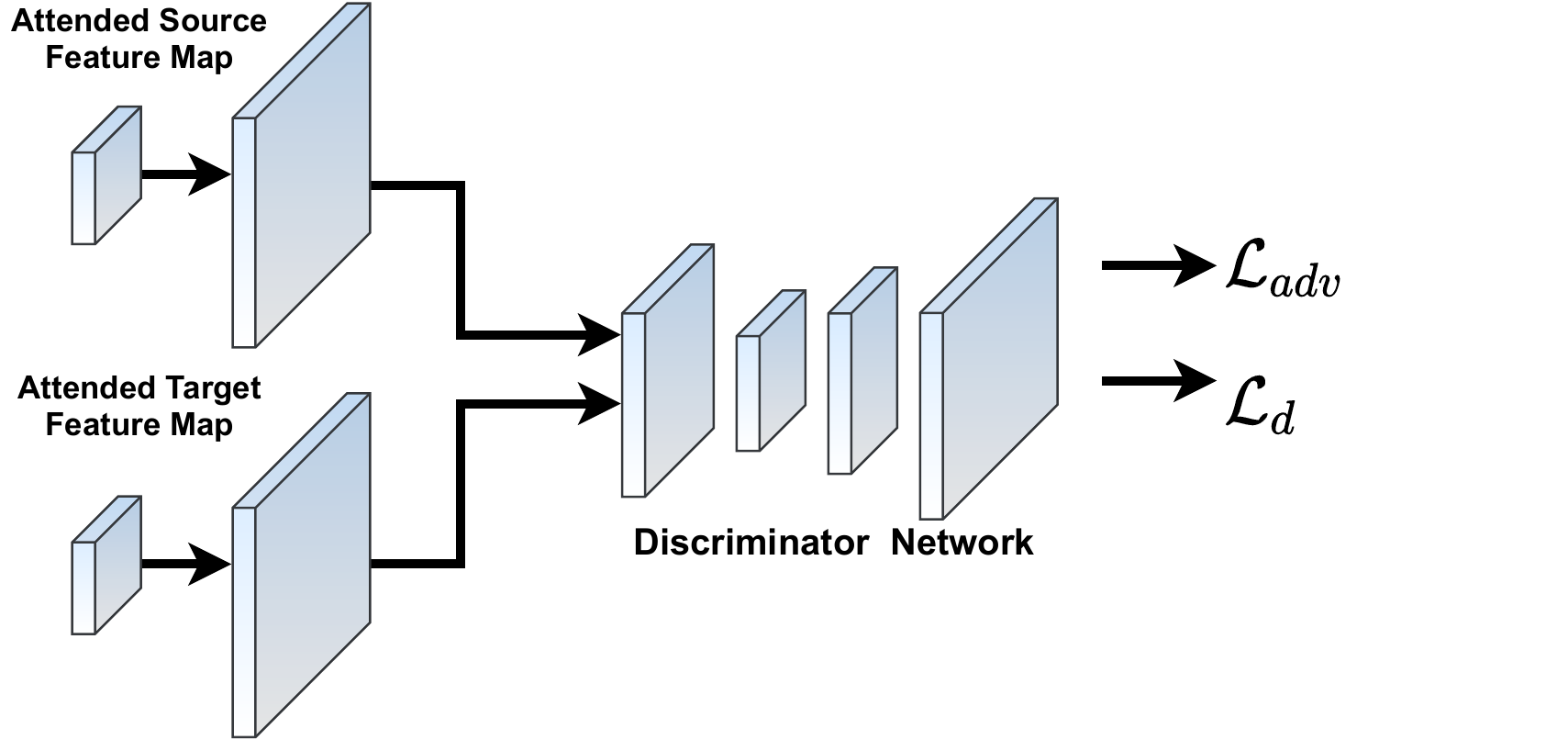}
  \caption{An overview of the ADAM module.}
  \vskip-3ex
  \label{fig.overview.attention.da}
\end{figure}

\textbf{Attentional domain adaptation module.}
Next, we design \emph{ADAM}, an \emph{attentional} domain adaptation module, aimed at detecting and  magnifying the significant amount of pinhole-panoramic correspondences at both, local and global levels (overview in Fig.~\ref{fig.overview.attention.da}).
ADAM differs from SDAM as it leverages the attention mechanism to learn an optimal weighting scheme for the  discriminator input. 
As in the Dual Attention Module~(DAM)~\cite{danet} shown in Fig.~\ref{fig.overview}, the feature map extracted by the backbone model is denoted as $F \in \mathbb{R}^{h \times w \times c}$, where the $h$, $w$, and $c$ are the height, width and channel of the feature map. After this representation is reshaped as $F' \in \mathbb{R}^{(h \times w) \times c}$, the position-wise attended feature is calculated as: $S = \sigma(F'^T, F'\otimes F'^T), S \in \mathbb{R}^{c \times (h \times w)}$, where $\sigma$ is the Softmax function.
Similarly, the channel-wise attended feature is denoted as $R = \sigma(F', F'^T\otimes F'), R \in \mathbb{R}^{(h \times w)\times c}$. Then, the final dual attended feature is concatenated with the the reshaped $S'\in \mathbb{R}^{h \times w \times c}$ and the reshaped $R'\in \mathbb{R}^{h \times w \times c}$.
By doing this, ADAM enables direct context information exchange among all pixels, mitigating the influence of discrepancy in positional priors and local distortions.
Relevant portions of the feature maps of both, $x_s$ and $x_t$ inputs are enhanced through the  attention and the \emph{re-weighted} source and target representations are both used to optimize the corresponding discriminator \textbf{D}.

\begin{figure}[!t]
  \centering
  \includegraphics[width=0.485\textwidth]{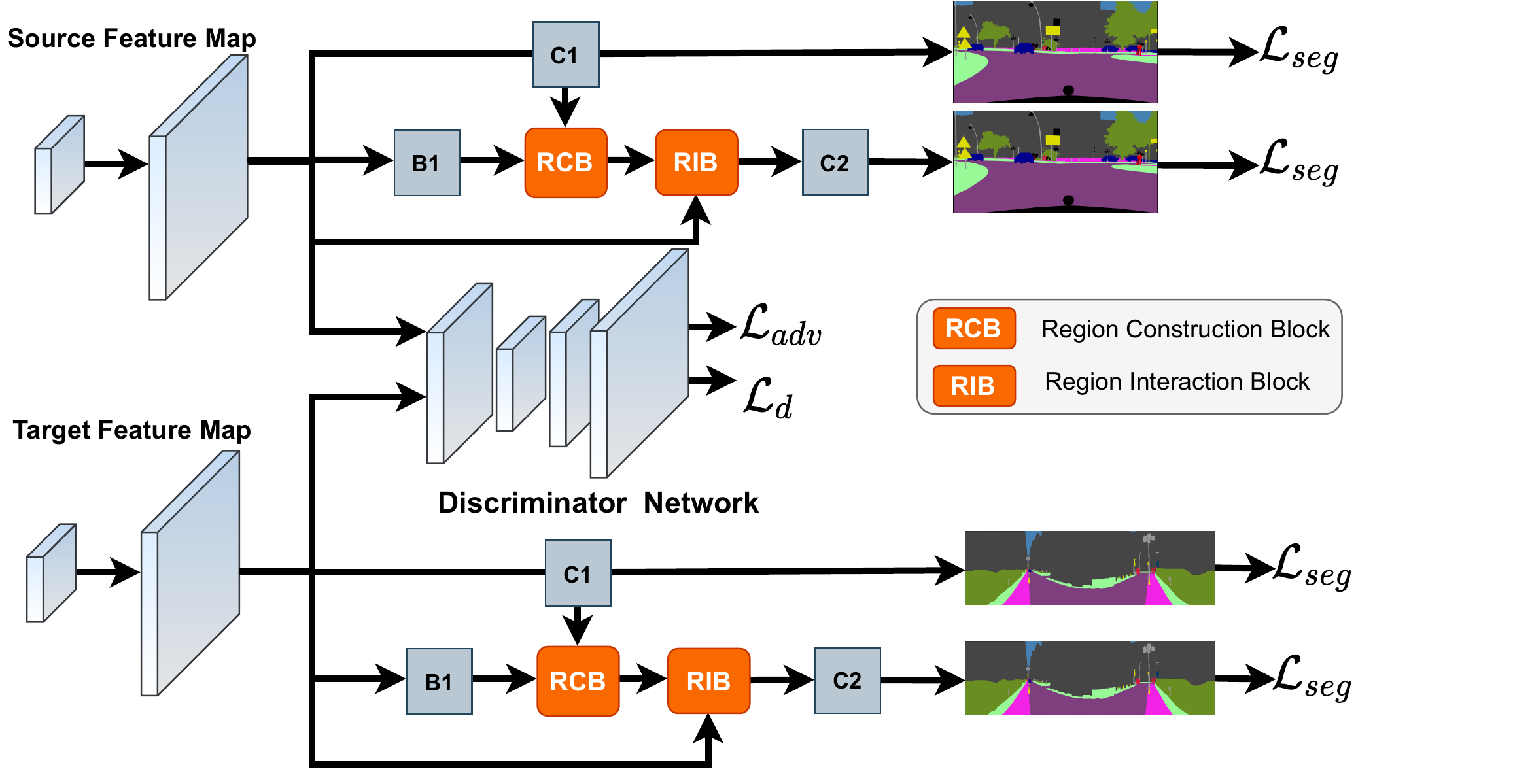}
  \caption{An overview of the RCDAM module.}
  \label{fig.overview.regional.da}
  \vskip-3ex
\end{figure}

\begin{figure}[!t]
  \centering
  \includegraphics[width=0.485\textwidth]{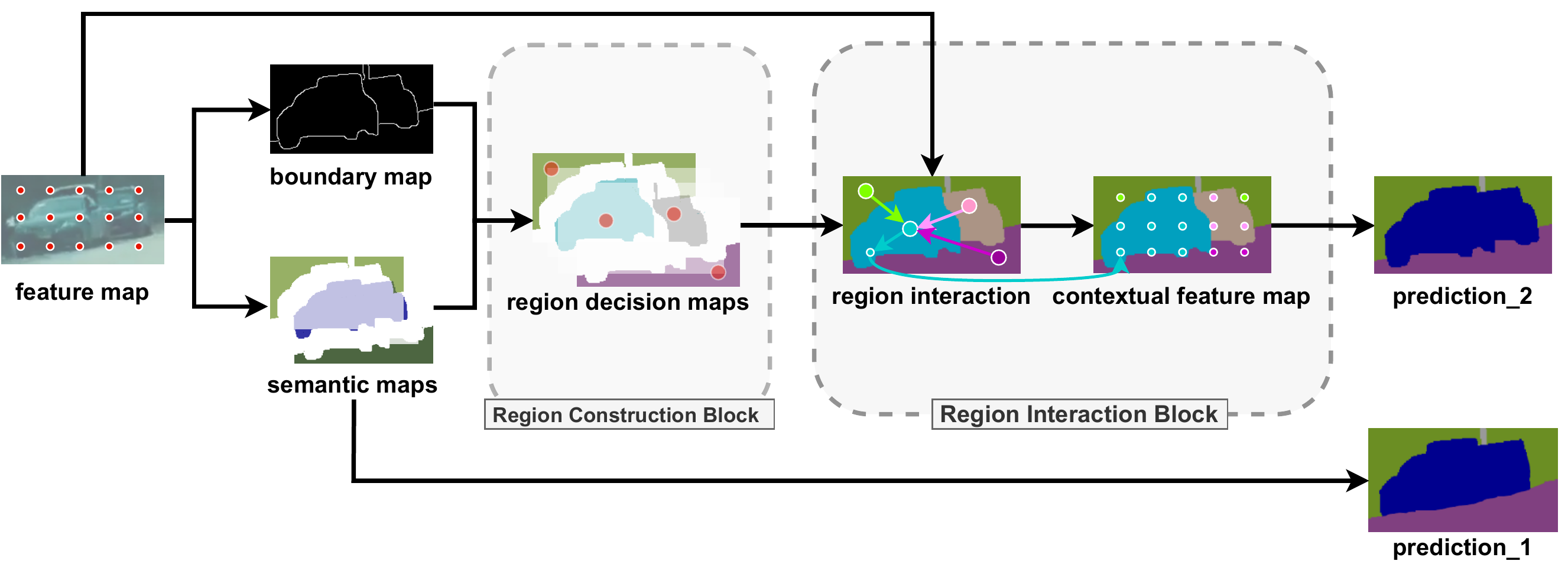}
  \caption{Diagram of Region Construction and Interaction Blocks.}
  \label{fig.overview.blocks.da}
  \vskip-3ex
\end{figure}

\textbf{Regional domain adaptation module.}
Next, we focus on \emph{region relationship of the panoramic images}.
Inspired by RANet, we design the RCDAM module based on the Regional Attention Module (RAM)~\cite{ranet} to configure the information flow between different regions and within the same region, as illustrated in Fig.~\ref{fig.overview.regional.da}. 
RCDAM follows a hierarchical adversarial learning scheme with two-stage discriminators, where the first stage is identical to the previously described SDAM.
The second stage is conducted by the RAM module, which includes two blocks: a Region Construction Block (RCB) and a Region Interaction Block (RIB) first introduced in RANet.
The inputs to this stage are the feature maps of $F_s$ and $F_t$ after a segmentation network forward pass.
Fig.~\ref{fig.overview.blocks.da} gives a detailed overview of the RCB and RIB building blocks. 

In contrast to~\cite{ranet}, we define three classifiers for the given feature map.
As shown in Fig.~\ref{fig.overview}, the \emph{B1} classifier outputs the binary boundary map, whereas the \emph{C1} and \emph{C2} classifiers are for the semantic maps according to the number of classes.
To perform the hierarchical learning and to lead the discrepancy between two semantic classifiers \emph{C1} and \emph{C2}, we utilize the semantic classifier from the original RCB block.
Then, RCB receives the binary boundary map and semantic maps as in~\cite{ranet} to link each pixel (the red dot in Fig.~\ref{fig.overview.blocks.da}) to different semantic regions, so as to conduct a region decision map with the representative pixels (the larger dots). 
Based on the region decision map, RIB can select representative pixels (the larger dots in color) for each region. According to the given feature map, RIB will back-project the prototypical information to each pixel (the smaller dots) in the same region to form the global contextual representations which are afterwards propagated to all pixels in the image.
Finally, regional context is sufficiently exchanged and the regional context-driven feature maps are built to generate final prediction outputs, which will be used to perform the same operations as in the first stage. 

It is worth noting that the second stage helps improving the segmentation result in the first stage, which can make  the outcome more compact and helps the \textsc{Pinhole$\rightarrow$Panoramic} adaptation  by exchanging regional context. 
Consequently, the given feature map belongs to the first stage of our framework (see Fig.~\ref{fig.overview.regional.da}) and this domain adaptation module does not impact the original segmentation architecture. It can therefore be flexibly used in models with or without attention layers.

\textbf{Feature confidence domain adaptation module.}
Compared to the aforementioned domain adaptation model, our next module FCDAM mainly operates in the \emph{feature confidence space}.
After the model undergoes the alignment operation in feature- and output-space, FCDAM is used to further improve the confidence of domain-specific features given by the backbone architecture.
Different from~\cite{advent}, the \emph{entropy map} $E \in [0, 1]^{h\times w}$ is calculated by the given feature map. Thus, the loss of entropy map is formulated as: $\mathcal{L}_{ent}(F) = -\sum_{h, w}(\phi(F^{(h, w)})\text{log}(\phi(F^{(h, w)})))$, where $\phi$ is the Sigmoid function applied at each pixel of feature map $F\in \mathbb{R}^{h\times w}$. During training \textbf{G} with the feature map $F_s=G(x_s)$ and $F_t=G(x_t)$ from source- and target domain, FCDAM can improve the feature confidence by minimizing the loss of the feature entropy map. Differing from previous DA modules, $\mathcal{L}_{adv}(G)$ for FCDAM in Eqn.~\eqref{eqn:finale_loss} is replaced by $\mathcal{L}_{ent}(G(F)) = \lambda_{ent}^s\mathcal{L}_{ent}(G(F_s)) + \lambda_{ent}^t\mathcal{L}_{ent}(G(F_t))$, where both $\lambda_{ent}$ are same as $\lambda_{adv}$. Note that, as shown in Fig.~\ref{fig.overview}, this FCDAM module eases the process of adding feature confidence learning to the original backbone without modification to the architecture of the whole domain adaptation framework.

\begin{figure*}[!t]
  \centering
  \vskip-3ex
  \includegraphics[width= \textwidth]{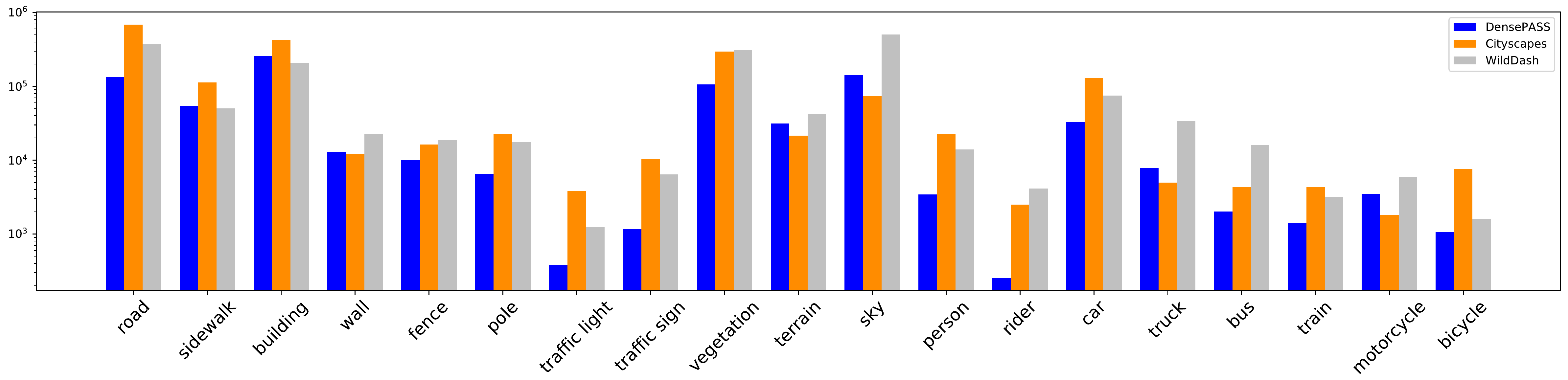}
  \caption{Distributions of DensePASS, Cityscapes~\cite{cityscapes}, and WildDash~\cite{wilddash} in terms of class-wise pixel counts per image.}
  \label{fig.distribution}
\vskip-1ex
\end{figure*}

\section{Experiments}
We conduct extensive experiments with different variants of our P2PDA framework in order to validate our the  idea  of  panoramic  segmentation through domain adaptation from the pinhole camera images.
First, we introduce \textsc{DensePASS} -- a novel  dataset for dense \emph{panoramic} segmentation of driving scenes annotated in accordance to the \emph{pinhole} camera  benchmark Cityscapes (Sec. \ref{sec:dataset}). 
Then, we  quantitatively evaluate how well our P2PDA framework can handle the \textsc{Pinhole$\rightarrow$Panoramic} transfer and conduct extensive ablation studies for different versions of DA modules and two  segmentation networks: the speed-oriented FANet~\cite{fanet} (Sec. \ref{sec:fanet}) and accuracy-oriented  DANet~\cite{danet} (Sec. \ref{sec:danet}).
We further benchmark our framework against $>20$ state-of-the-art semantic segmentation models (Sec. \ref{sec:soa}), compare our approach with competitive panoramic segmentation and domain adaptation frameworks (Sec. \ref{sec:more}), and examine the impact of expanding the training set  with more examples (Sec. \ref{sec:wilddash}).
Finally, we showcase multiple qualitative results (Sec. \ref{sec:qualitative}).
We adopt Mean Intersection over Union (mIoU) as our primary evaluation metric.

\subsection{Datasets and Experimental Settings}
\label{sec:dataset}

\textbf{Source dataset (pinhole).} 
We leverage  Cityscapes~\cite{cityscapes} as our label-rich source dataset providing a high amount of annotated pinhole camera images.
Cityscapes comprises $2979$ training- and $500$ validation images ($1024{\times}2048$ resolution) captured in $50$ different European cities and annotated with $19$ categories. 
In most experiments, we use the $2979$ training samples as our source training data.
To investigate the potential of P2PDA, in some cases, we also leverage the WildDash~\cite{wilddash} dataset which includes $4256$ pinhole images ($1080{\times}1920$ resolution) for improving the generalization.   

\textbf{DensePASS target dataset (panoramic).} 
Since no established segmentation benchmarks address the \textsc{Pinhole$\rightarrow$Panoramic} recognition and previous panoramic test-beds cover a very limited number of classes~\cite{pass}\cite{omnirange}, we collect DensePASS -- a novel densely annotated dataset for panoramic segmentation of driving scenes.
It is created with the \textsc{Pinhole$\rightarrow$Panoramic} transfer in mind, so that the test data is annotated with $19$ categories present in the pinhole camera dataset Cityscapes and other prominent semantic segmentation benchmarks~\cite{wilddash,bdd}.
To facilitate the unsupervised domain adaptation task, DensePASS covers both, labelled data ($100$ panoramic images used for testing) and unlabelled training data ($2000$ panoramic images used for the domain transfer optimization).
A FoV of $70^\circ{\times}360^\circ$ is covered in the captured panoramic images with a $400{\times}2048$ resolution.
The data is collected using Google Street View and includes images from different continents ($25$ different cities for testing and $40$ for training).

In Fig.~\ref{fig.distribution}, we compare label distributions of DensePASS with Cityscapes~\cite{cityscapes} and WildDash~\cite{wilddash} datasets in terms of pixel counts by averaging the number of images used in our domain adaptation study.
Our histogram analysis indicates, that DensePASS and the mentioned pinhole camera datasets follow a relatively close distribution of categories.
This observation indicates, that distribution alignment not only in the feature-space but also in the semantic output-space might be beneficial and is therefore integrated in our framework.

\textbf{Training settings.}
We use stochastic gradient descent (initial learning rate of $1e-5$, momentum of $0.9$, decay of $5e-4$) for optimization of the segmentation network \textbf{G} and the Adam optimizer~\cite{adam} for  discriminators \textbf{D} (initial learning rate set to $4e-6$).
For both optimizers, the learning rate is decreased  polynomially through multiplication with  $(1-\frac{iter}{max\_iter})^{0.9}$ after each iteration, where max\_iter is set to $200000$ with a batch-size of $2$. The semantic loss (\emph{i.e.} the cross-entropy loss) in Eqn.~\eqref{eqn:seg_loss} is updated with class weights, which are calculated in advance from the source annotated dataset, following ERFNet~\cite{erfnet}.
The loss balancing weights regarding to Eqn.~\eqref{eqn:finale_loss}, $\lambda_{adv}$ and $\lambda_{seg}$ are set to $0.001$ and $1.0$ for SDAM/FCDAM, and $0.0002$ and $0.1$ for ADAM.
In the RCDAM module, $\lambda_{seg}$ is set to $1.5$ for the prediction result before RCB. 
During training, pinhole data is resized to $720{\times}1280$ while panoramic images remain at the $400{\times}2048$ resolution.
In the FCDAM training stage and the pseudo-label self-supervised learning stage, the learning rate is decreased as $1e-8$ for the generators and $4e-9$ for the discriminators.
Horizontal flipping and random translation between $[-2, 2]$ pixels are performed for data augmentation.

\begin{table*}[!t]
\renewcommand\arraystretch{1.2}
      \footnotesize
      \setlength{\tabcolsep}{4pt}
      \begin{center}
      \caption{Per-class results on DensePASS. We use FANet~\cite{fanet} as the segmentation network and set different domain adaptation modules on our P2PDA framework to test on DensePASS with the size of input $2048{\times}400$. S, A, R, F represent the SDAM, ADAM, RCDAM and FCDAM respectively. Feature- and output-space are named as FS and OS for short. SSL represents the self-supervised learning with pseudo-labels. The first line is the Cityscapes source-only result without adaptation. }
      \label{tab:fanet}
      \begin{adjustbox}{max width=\textwidth}
          \begin{tabular}{ l | c |c | c | c c c c c c c c c c c c c c c c c c c}
              \toprule[1pt]
         Methods & FS & OS &  \rotatebox{90}{Mean IoU} &  \rotatebox{90}{road} &  \rotatebox{90}{sidewalk} &  \rotatebox{90}{building} & \rotatebox{90}{ wall} &  \rotatebox{90}{fence} &  \rotatebox{90}{pole} & \rotatebox{90}{traffic light} &  \rotatebox{90}{traffic sign}&  \rotatebox{90}{vegetation} &  \rotatebox{90}{terrain} &  \rotatebox{90}{sky} & \rotatebox{90}{person} &  \rotatebox{90}{rider} & \rotatebox{90}{car} &  \rotatebox{90}{truck}& \rotatebox{90}{ bus}& \rotatebox{90}{ train}& \rotatebox{90}{ motorcycle}&  \rotatebox{90}{bicycle}\\
        \hline
        \hline
        FANet & - & - & 26.90 & \textbf{62.98} & 10.64 & 72.41 & 7.80 & 20.74 & 11.77 & 6.85 & 3.75 & 68.11 & 21.56 & 87.00 & 23.73 & 5.33 & 49.61 & 10.65 & \textbf{0.54} & 16.76 & 24.15 & 6.62 \\
        FANet & S & S & 32.17 & 62.16 & 16.85 & 78.78 & 13.67 & 24.07 & 19.72 & 11.42 & 9.68 & 71.42 & 18.22 & 85.72 & 32.66 & \textbf{11.75} & 54.34 & 17.61 & 0.00 & 41.52 & 29.30 & 12.30 \\
        FANet & A & S & 32.67 & 62.28 & 16.86 & 79.99 & \textbf{17.64} & 23.96 & \textbf{19.78} & \textbf{12.33} & 9.58 & 72.01 & 19.29 & 85.91 & 32.85 & 11.03 & 55.75 & 15.38 & 0.38 & 43.53 & 29.19 & 12.95 \\
        FANet & S+A & S & 33.05 & 61.74 & 17.70 & \textbf{80.07} & 16.38 & 24.64 & 19.61 & 12.04 & \textbf{9.79} & \textbf{72.27} & 17.94 & 86.31 & 33.17 & 11.47 & 55.18 & 15.61 & 0.04 & 52.55 & 28.68 & 12.82 \\
        FANet & S+A & R & 33.02 & 62.58 & 19.25 & 80.07 & 15.68 & 24.87 & 19.27 & 11.54 & 9.01 & 71.95 & 19.65 & 86.89 & 32.18 & 12.03 & 55.12 & 17.37 & 0.21 & 44.98 & \textbf{29.93} & 14.87 \\
        FANet & S+A+F & R & 33.52 & 57.16 & 25.66 & 78.43 & 16.02 & \textbf{26.88} & 12.76 & 2.30 & 7.34 & 68.73 & 26.92 & 87.45 & \textbf{36.51} & 1.20 & 62.83 & 20.16 & 0.00 & 68.46 & 17.86 & 20.19 \\
        FANet-SSL & S+A & R & 34.26 & 57.92 & 24.22 & 78.84 & 14.94 & 25.42 & 13.39 & 4.82 & 7.14 & 69.47 & 25.77 & \textbf{87.92} & 36.12 & 4.27 & 62.83 & 22.90 & 0.00 & 78.73 & 16.15 & 20.02 \\
        FANet-SSL & S+A+F & R & \textbf{35.67} & 58.08 & \textbf{28.75} & 78.19 & 16.47 & 26.86 & 13.78 & 4.76 & 7.62 & 69.01 & \textbf{34.58} & 87.51 & 36.12 & 0.90 & \textbf{64.06} & \textbf{27.50} & 0.00 & \textbf{84.99} & 18.13 & \textbf{20.35} \\
          \bottomrule[1pt]
          \end{tabular}
      \end{adjustbox}
      \end{center}
  \end{table*}

Different from the attended feature map setting of DANet, which forwards the attended feature map with a downsampling rate of $16$ as the input of two classifiers (\emph{B1} and \emph{C1}), FANet has multi-level attended feature maps. After our experiment, the feature maps with downsampling rates of $16$ and $4$ are concatenated and upsampled as input to the \emph{B1} classifier. At the same time, feature maps with downsampling rates of $16$ and $8$ are concatenated and upsampled as input to the \emph{C1} classifier in the FANet setting.

\subsection{Ablation Studies for Segmentation Network with FANet}
\label{sec:fanet}

We first consider FANet~\cite{fanet}, a lightweight speed-oriented network, as the segmentation model and investigate different combinations of our four domain adaptation modules. 
As shown in Table~\ref{tab:fanet}, before adaptation, FANet yields a mIoU of $26.90\%$ indicating large room for improvement in cross-domain generalization.
Our framework improves the result to $32.17\%$ by using the SDAM module in both feature- and output-space ($+5.27\%$ gain).
Integrating the attentional ADAM module also leads to a considerable boost ($32.67\%$ in mIoU, a $+5.77\%$ gain over the source-only baseline). 
A combination of our four modules yields the recognition result of $33.52\%$ in mIoU. Furthermore, the pseudo-label self-supervised learning boosts our S+A+R and S+A+F+R adaptation results to $34.26\%$ and $35.67\%$ in mIoU, respectively.

\begin{table*}[!t]
\renewcommand\arraystretch{1.2}
      \footnotesize
      \setlength{\tabcolsep}{4pt}
      \begin{center}
      \caption{Per-class results on DensePASS. We use DANet~\cite{danet} as the segmentation network and set different domain adaptation modules on our P2PDA framework to test on DensePASS with the size of input $2048{\times}400$. S, A, R, F represent the SDAM, ADAM, RCDAM and FCDAM respectively. Feature- and output-space are named as FS and OS for short. SSL represents the self-supervised learning with pseudo-labels. The first line is the Cityscapes source-only result without adaptation. * denotes further adding source images from WildDash to complement Cityscapes.}
      \label{tab:danet}
      \begin{adjustbox}{max width=\textwidth}
          \begin{tabular}{ l | c |c | c | c c c c c c c c c c c c c c c c c c c}
              \toprule[1pt]
         Methods & FS & OS &  \rotatebox{90}{Mean IoU} &  \rotatebox{90}{road} &  \rotatebox{90}{sidewalk} &  \rotatebox{90}{building} & \rotatebox{90}{ wall} &  \rotatebox{90}{fence} &  \rotatebox{90}{pole} & \rotatebox{90}{traffic light} &  \rotatebox{90}{traffic sign}&  \rotatebox{90}{vegetation} &  \rotatebox{90}{terrain} &  \rotatebox{90}{sky} & \rotatebox{90}{person} &  \rotatebox{90}{rider} & \rotatebox{90}{car} &  \rotatebox{90}{truck}& \rotatebox{90}{ bus}& \rotatebox{90}{ train}& \rotatebox{90}{ motorcycle}&  \rotatebox{90}{bicycle}\\
        \hline
        \hline
            DANet & - & - & 28.50 & \textbf{70.68} & 8.30 & 75.80 & 9.49 & 21.64 & \textbf{15.91} & 5.85 & 9.26 & \textbf{71.08} & 31.50 & 85.13 & 6.55 & 1.68 & 55.48 & 24.91 & 30.22 & 0.52 & 0.53 & 17.00 \\
            DANet & S & S & 38.51 & 61.78 & 21.11 & 74.59 & 22.59 & 29.93 & 14.79 & 15.00 & 10.17 & 66.94 & 19.03 & 82.57 & 31.03 & 21.24 & 53.26 & 54.67 & 37.77 & 39.40 & 43.84 & 31.95 \\
            DANet & A & S & 39.16 & 61.34 & 20.71 & 76.52 & 20.53 & 30.03 & 14.19 & 15.69 & 10.09 & 68.60 & 18.84 & 82.08 & 33.16 & 21.75 & 57.68 & 53.88 & 40.33 & 41.47 & 46.11 & 31.00 \\
            DANet & S+A & S & 39.28 & 62.43 & 21.89 & 76.22 & 21.42 & 30.54 & 14.85 & 14.10 & 9.76 & 69.07 & 19.94 & 82.84 & \textbf{34.56} & 19.30 & 56.51 & 53.04 & 42.51 & 39.47 & 45.71 & 32.09 \\
            DANet & S & R & 39.46 & 62.75 & 23.17 & 76.65 & 23.90 & \textbf{30.82} & 14.84 & \textbf{18.44} & 10.09 & 69.10 & 17.60 & 82.78 & 33.51 & 21.53 & 55.97 & 51.78 & 41.77 & 36.90 & 46.11 & \textbf{32.12} \\
            DANet & S+A & R & 39.76 & 63.11 & 24.63 & 76.17 & 25.03 & 30.56 & 13.68 & 15.68 & \textbf{10.53} & 67.31 & 22.41 & 80.15 & 32.95 & 21.11 & 54.39 & 53.51 & 43.64 & 42.20 & 46.71 & 31.66 \\
            DANet & S+A+F & R & 40.52 & 62.90 & 25.58 & 76.62 & 24.45 & 30.37 & 14.45 & 16.75 & 9.96 & 67.87 & 19.70 & 82.04 & 34.18 & \textbf{22.95} & 56.99 & 54.27 & \textbf{44.15} & 47.75 & 46.98 & 31.86  \\
            DANet-SSL & S+A & R & 41.39 & 67.24 & 27.98 & 77.18 & 25.11 & 25.80 & 15.33 & 10.59 & 6.58 & 69.24 & 33.89 & 80.96 & 32.18 & 5.29 & 69.86 & 59.70 & 36.20 & 65.99 &  \textbf{47.47} & 29.87\\
            DANet-SSL & S+A+F & R & \textbf{41.99} & 70.21 & \textbf{30.24} & \textbf{78.44} & \textbf{26.72} & 28.44 & 14.02 & 11.67 & 5.79 & 68.54 & \textbf{38.20} & \textbf{85.97} & 28.14 & 0.00 & \textbf{70.36} & \textbf{60.49} & 38.90 & \textbf{77.80} & 39.85 & 24.02  \\
            
            \midrule
            DANet* & S & R & 41.35 & 68.38 & \textbf{37.26} & 75.51 & 26.28 & 31.81 & 15.62 & 8.99 & 10.33 & 66.22 & 31.74 & 80.68 & 33.69 & 16.81 & 64.81 & 47.67 & 28.05 & 61.81 & 44.92 & 34.98 \\
            DANet* & S+A & R & 42.47 & 67.47 & 30.16 & 75.27 & 30.26 & 37.50 & 16.19 & 9.35 & 9.78 & 63.14 & 30.44 & 77.07 & 34.82 & 15.24 & 64.33 & 53.70 & 43.33 & 71.57 & 46.80 & 30.47 \\
            DANet* & S+A+F & R & 42.87 & 66.92 & 29.97 & 77.34 & \textbf{30.87} & \textbf{37.85} & 15.04 & 11.12 & 9.60 & 62.80 & 31.03 & 78.08 & 36.27 & 18.01 & 63.66 & 54.83 & 42.86 & 74.22 & 45.96 & 28.13 \\
            DANet-SSL* & S+A & R & 44.27 & 70.63 & 35.30 & 78.52 & 25.27 & 33.51 & 14.43 & \textbf{13.80} & 7.31 & 63.52 & 34.94 & 84.31 & 34.54 & \textbf{19.08} & 70.05 & 49.14 & \textbf{48.80} & \textbf{75.11} & 47.53 & \textbf{35.36} \\
            DANet-SSL* & S+A+F & R & \textbf{44.66} & \textbf{75.85} & 34.21 & \textbf{82.58} & 28.75 & 35.58 & \textbf{18.51} & 12.65 & \textbf{12.49} & \textbf{71.33} & \textbf{37.51} & \textbf{89.80} & \textbf{38.68} & 15.99 & \textbf{76.59} & \textbf{62.81} & 12.25 & 61.56 & \textbf{48.18} & 33.26 \\
          \bottomrule[1pt]
          \end{tabular}
      \end{adjustbox}
      \end{center}
  \end{table*}
  
\subsection{Ablation Studies for Segmentation Network with DANet}
\label{sec:danet}  
  
Our main architecture for the in-depth experiments is the accuracy-oriented segmentation network DANet~\cite{danet} (results provided in Table~\ref{tab:danet}).
The native source-trained DANet achieves a mIoU of only $28.50\%$, highlighting the sensitivity of modern segmentation networks to the \textsc{Pinhole$\rightarrow$Panoramic} domain shift.
The performance is strongly improved ($+10.01\%$ boost) through SDAM modules placed in feature- and output-space, achieving $38.51\%$ in mIoU.
Similarly, using the ADAM module yields a result of $39.16\%$ (a $+10.66\%$ improvement over the source-only baseline). 
Combining both the SDAM and ADAM modules again slightly improves the performance ($39.28\%$ in mIoU).

We further explore the use of the RCDAM module in output-space, yielding  $39.46\%$ mIoU ($+10.96\%$ boost over the baseline).
The performance of $39.76\%$ ($+11.26\%$ boost with respect to the original segmentation network) is achieved by combining three modules: SDAM, ADAM and RCDAM.
Integrating FCDAM leads $40.52\%$ in mIoU (a $+12.02\%$ increase).
Overall, our experiments showcase that  direct cross-domain semantic segmentation is a hard task and P2PDA framework with attentional and regional domain adaptation modules clearly helps to close the domain gap.
Furthermore, applying pseudo-label self-supervised learning based on prediction uncertainties improves the results to $41.99\%$ in mIoU.

\begin{table}[!t]
\caption{Performance of CNN- and transformer-based semantic segmentation models on Cityscapes and DensePASS.}
\label{tab:domain_gap}
\centering
\resizebox{\columnwidth}{!}{
\begin{tabular}{@{}llrrc@{}}
\toprule
\textbf{Network} & \textbf{Backbone} & \textbf{Cityscapes} & \textbf{DensePASS} & \textbf{mIoU Gap} \\ \midrule \midrule
SwiftNet~\cite{swiftnet} & ResNet-18 & 75.4 & 25.7 & -49.7\\
DeepLabV3+~\cite{deeplabv3+} & ResNet-18    & 76.8  &  25.6   & -51.2 \\ 
OCRNet~\cite{ocrnet}            & HRNetV2p-W18s & 77.1 & 25.9   & -51.2 \\
Fast-SCNN~\cite{fastscnn}     & Fast-SCNN	   & 69.1  &  24.6   & -44.5 \\
\midrule
DeepLabV3+~\cite{deeplabv3+} & ResNet-50    & 80.1  &  29.0  & -51.1 \\ 
PSPNet~\cite{pspnet}            & ResNet-50    & 78.6  &  29.5  & -49.1 \\
DNL~\cite{dnl}                   & ResNet-50    & 79.3  & 28.7  & -50.6 \\
Semantic-FPN~\cite{panopticfpn} & ResNet-50 & 74.5 &  29.9   & -44.6 \\
OCRNet~\cite{ocrnet}            & HRNetV2p-W18 & 78.6  & 30.8  & -47.8 \\ 
\midrule
DeepLabV3+~\cite{deeplabv3+} & ResNet-101   & 80.9  &  32.5  & -48.4 \\
PSPNet~\cite{pspnet}            & ResNet-101   & 79.8  &  30.4  & -49.4 \\
DANet~\cite{danet}                 & ResNet-101   & 80.4  &  28.5   & -51.9 \\
DNL~\cite{dnl}                   & ResNet-101   & 80.4  & 32.1  & -48.3 \\
Semantic-FPN~\cite{panopticfpn} & ResNet-101 & 75.8 &  28.8  & -47.0 \\
ResNeSt~\cite{resnest}         & ResNeSt-101  & 79.6  &  28.8   & -50.8 \\
OCRNet~\cite{ocrnet}            & HRNetV2p-W48 & 80.7  &  32.8  & -47.9 \\
\midrule
SETR-MLA~\cite{setr} & Transformer-Large & 77.2 & 35.6 & -41.6 \\
SETR-PUP~\cite{setr} & Transformer-Large & 79.3 & 35.7 & -43.6 \\
\midrule \midrule
ERFNet~\cite{erfnet} & ERFNet    & 72.1  & 16.7 &  -55.4  \\ 
ERFNet~\cite{erfnet} (Ours) & ERFNet & 72.1 & 34.1 & \textbf{-38.0} \\
FANet~\cite{fanet} & ResNet-34 & 71.3 & 26.9 & -44.4 \\
FANet~\cite{fanet} (Ours) & ResNet-34 & 71.3 & 35.7 & \textbf{-35.6} \\
DANet~\cite{danet} & ResNet-50 & 79.3  &  28.5  & -50.8 \\
DANet~\cite{danet} (Ours) & ResNet-50 & 79.3 & 42.0 & \textbf{-37.3} \\
\bottomrule
\end{tabular}}
\end{table}

\subsection{Benchmarking and Comparison with the State-of-the-Art}
\label{sec:soa}

Until now, we compared different framework configurations with each other and the native segmentation network.
Next, we aim to quantify pinhole-panoramic domain gap and extend our evaluation with over $20$ off-the-shelf segmentation models trained on Cityscapes and evaluated on both, Cityscapes (no domain shift) and the panoramic DensePASS images.\footnote[1]{For a fair comparison, model weights are provided by the same framework MMSegmentation: https://github.com/open-mmlab/mmsegmentation}
Table~\ref{tab:domain_gap} summarizes our results.
It is evident, that modern CNN-based segmentation models trained on pinhole camera images struggle with generalization to  panoramic data, with performance degrading by $\sim50\%$ as we move from the standard \textsc{Pinhole$\rightarrow$Pinhole} setting to the \textsc{Pinhole$\rightarrow$Panoramic} evaluation. 
The recent transformer-based method SETR~\cite{setr} with a powerful computation-intensive backbone is more robust, yet also suffering from $>40\%$ accuracy drops.

We also verify our P2PDA domain adaptation strategy with three different segmentation models (bottom part of Table~\ref{tab:domain_gap}).
Our P2PDA strategy with regional and attentional context exchange improves the \textsc{Pinhole$\rightarrow$Panoramic} outcome by a large margin (mIoU gains of $+17.4\%$, $+8.8\%$ and $+13.5\%$ for ERFNet, FANet and DANet, respectively).
Our experiments provide encouraging evidence that P2PDA can be successfully deployed for cross-domain $360^\circ$ understanding. 

We now consider the inference speed and test the forward pass time with the batch-size of $1$ to stimulate real driving applications. We report the mean Frames Per Second (FPS) running through the $100$ panoramic images at the resolution of $400{\times}2048$ on a GTX 1080Ti GPU processor. It turns out that ERFNet, FANet and DANet reach $32.3$, $67.7$ and $17.2$ FPS, respectively. We see clear speed-accuracy trade-offs, with FANet being the fastest and DANet being the most accurate.

\begin{table*}[!t]
\renewcommand\arraystretch{1.2}
      \footnotesize
      \setlength{\tabcolsep}{4pt}
      \begin{center}
      \caption{Per-class results on DensePASS. Comparison with state-of-the-art panoramic semantic segmentation (PASS~\cite{pass} and ECANet~\cite{omnirange}), unsupervised domain adaptation (CLAN~\cite{clan} and CRST~\cite{crst}), and multi-supervision methods~\cite{issafe,seamless,usss}. * denotes performing two runs of the self-supervised learning.}
      \label{tab:more}
      \begin{adjustbox}{max width=\textwidth}
          \begin{tabular}{ l | c | c c c c c c c c c c c c c c c c c c c}
              \toprule[1pt]
         Methods & \rotatebox{90}{Mean IoU} &  \rotatebox{90}{road} &  \rotatebox{90}{sidewalk} &  \rotatebox{90}{building} & \rotatebox{90}{ wall} &  \rotatebox{90}{fence} &  \rotatebox{90}{pole} & \rotatebox{90}{traffic light} &  \rotatebox{90}{traffic sign}&  \rotatebox{90}{vegetation} &  \rotatebox{90}{terrain} &  \rotatebox{90}{sky} & \rotatebox{90}{person} &  \rotatebox{90}{rider} & \rotatebox{90}{car} &  \rotatebox{90}{truck}& \rotatebox{90}{ bus}& \rotatebox{90}{ train}& \rotatebox{90}{ motorcycle}&  \rotatebox{90}{bicycle}\\
        \hline
        \hline
        ERFNet~\cite{erfnet} & 16.65 & 63.59 & 18.22 & 47.01 & 9.45 & 12.79 & 17.00 & 8.12 & 6.41 & 34.24 & 10.15 & 18.43 & 4.96 & 2.31 & 46.03 & 3.19 & 0.59 & 0.00 & 8.30 & 5.55 \\
        PASS~\cite{pass} (ERFNet) & 23.66 & 67.84 & 28.75 & 59.69 & 19.96 & 29.41 & 8.26 & 4.54 & 8.07 & 64.96 & 13.75 & 33.50 & 12.87 & 3.17 & 48.26 & 2.17 & 0.82 & 0.29 & 23.76& 19.46 \\
        \midrule
        ECANet (Omni-supervised)~\cite{omnirange} & 43.02 & \textbf{81.60} & 19.46 & 81.00 & 32.02 & \textbf{39.47} & \textbf{25.54} & 3.85 & 17.38 & \textbf{79.01} & 39.75 & \textbf{94.60} & \textbf{46.39} & 12.98 & \textbf{81.96} & 49.25 & 28.29 & 0.00 & \textbf{55.36} & 29.47 \\
        \midrule
        CLAN (Adversarial training)~\cite{clan} & 31.46 & 65.39 & 21.14 & 69.10 & 17.29 & 25.49 & 11.17 & 3.14 & 7.61 & 71.03 & 28.19 & 55.55 & 18.86 & 2.76 & 71.60 & 26.42 & 17.99 & 59.53 & 9.44 & 15.91\\
        CRST-LRENT (Self-training)~\cite{crst} & 31.67 & 68.18 & 15.72 & 76.78 & 14.06 & 26.11 & 9.90 & 0.82 & 2.66 & 69.36 & 21.95 & 80.06 & 9.71 & 1.25 & 65.12 & 38.76 & 27.22 & 48.85 & 7.10 & 18.08\\
        \midrule
        Seamless (Mapillary)~\cite{seamless} & 34.14 & 59.26 & 24.48 & 77.35 & 12.82 & 30.91 & 12.63 & \textbf{15.89} & \textbf{17.73} & 75.61 & 33.30 & 87.30 & 19.69 & 4.59 & 63.94 & 25.81 & \textbf{57.16} & 0.00 & 11.59 & 19.04 \\
        USSS (IDD)~\cite{usss} & 26.98 & 68.85 & 5.41 & 67.39 & 15.10 & 21.79 & 13.18 & 0.12 & 7.73 & 70.27 & 8.84 & 85.53 & 22.05 & 1.71 & 58.69 & 16.41 & 12.01 & 0.00 & 23.58 & 13.90 \\
        \midrule
        SwiftNet (ApolloScape) & 14.08 & 61.21 & 34.93 & 57.92 & 7.85 & 23.37 & 13.33 & 9.04 & 6.44 & 50.39 & 0.00 & 0.00 & 0.44 & 0.00 & 0.09 & 0.36 & 1.83 & 0.00 & 0.04 & 0.24 \\
        SwiftNet (Cityscapes)~\cite{swiftnet} & 25.67 & 50.73 & 32.76 & 70.24 & 12.63 & 24.02 & 18.79 & 7.18 & 4.01 & 64.93 & 23.70 & 84.29 & 14.91 & 0.97 & 43.46 & 8.92 & 0.04 & 4.45 & 12.77 & 8.77 \\
        SwiftNet (KITTI-360) & 25.00 & 69.03 & 27.71 & 68.07 & 15.70 & 16.26 & 15.29 & 0.00 & 4.43 & 64.71 & 31.01 & 84.86 & 23.02 & 0.00 & 45.08 & 9.72 & 0.00 & 0.00 & 0.00 & 0.00 \\
        SwiftNet (BDD) & 24.69 & 4.26 & 25.11 & 74.16 & 15.53 & 22.74 & 11.70 & 0.00 & 10.58 & 70.86 & 26.55 & 92.26 & 25.12 & 0.00 & 58.78 & 31.35 & 0.00 & 0.00 & 0.00 & 0.00 \\
        SwiftNet (Merge3)~\cite{issafe} & 32.04 & 68.31 & \textbf{38.59} & 81.48 & 15.65 & 23.91 & 20.74 & 5.95 & 0.00 & 70.64 & 25.09 & 90.93 & 32.66 & 0.00 & 66.91 & 42.30 & 5.97 & 0.07 & 6.85 & 12.66 \\
        \midrule
        Ours (Cityscapes) & 41.99 & 70.21 & 30.24 & 78.44 & 26.72 & 28.44 & 14.02 & 11.67 & 5.79 & 68.54 & 38.20 & 85.97 & 28.14 & 0.00 & 70.36 & 60.49 & 38.90 & 77.80 & 39.85 & 24.02\\
        Ours (Cityscapes+WildDash) & 44.66 & 75.85 & 34.21 & 82.58 & 28.75 & 35.58 & 18.51 & 12.65 & 12.49 & 71.33 & 37.51 & 89.80 & 38.68 & \textbf{15.99} & 76.59 & \textbf{62.81} & 12.25 & 61.56 & 48.18 & 33.26\\
        Ours* (Cityscapes+WildDash) & \textbf{48.52} & 76.87 & 35.70 & \textbf{85.16} & \textbf{33.93} & 38.86 & 18.18 & 10.52 & 13.71 & 73.98 & \textbf{41.89} & 92.08 & 42.38 & 8.26 & 78.62 & 60.12 & 42.17 & \textbf{81.21} & 53.82 & \textbf{34.49}\\
          \bottomrule[1pt]
          \end{tabular}
      \end{adjustbox}
      \end{center}
  \end{table*}

\subsection{Comparison to Panoramic Semantic Segmentation, Unsupervised Domain Adaptation, and Multi-Supervision Methods}
\label{sec:more}
Before delving into more comparisons, we note that the category definitions of ApolloScape~\cite{apolloscape}, IDD~\cite{idd}, and Mapillary Vistas~\cite{mapillary} datasets are different from other datasets, in which the identical $19$ categories following Cityscapes~\cite{cityscapes} can be obtained by class mapping. Models trained on ApolloScape~\cite{apolloscape} perform segmentation with $16$ overlapping categories, where the \emph{terrain}, \emph{sky}, and \emph{train} classes are discarded. The \emph{train} class in IDD~\cite{idd} and Mapillary Vistas~\cite{mapillary} is excluded, thus other $18$ classes are remained.

Next, we compare our approach with previous  segmentation frameworks specifically developed for panoramic images, including PASS~\cite{pass} and ECANet~\cite{omnirange}.
PASS elevates the accuracy of ERFNet by  fusing the semantically-meaningful features of panorama segments ($4$ segments as suggested by~\cite{pass}).
ECANet~\cite{omnirange} highlights horizontal dependencies and relies on omni-supervised learning using heavy training data, thereby reaching a high performance.
Our P2PDA-driven DANet trained with Cityscapes and WildDash sources (detailed in Sec.~\ref{sec:wilddash}) outperforms these works, achieving $44.66\%$ by using  pinhole data annotations only and successfully transferring beyond the FoV.
Performing another run of the self-supervised learning stage elevates the mIoU to $48.52\%$, leading to the best segmentation result.

We now compare P2PDA with two state-of-the-art approaches for unsupervised domain adaptation: one method based on adversarial learning (CLAN~\cite{clan}) and one built on self-training (CRST~\cite{crst}), both adapting from Cityscapes to DensePASS.
Our proposed framework clearly stands out in front of other domain adaptation pipelines, improving the performance by $\sim10\%$, showcasing the effectiveness of the attention-based design which intertwines attention-augmented adversarial learning and attention-regulated self-training for an effective knowledge transfer.

\begin{figure}[!t]
  \centering
  \includegraphics[width=0.485\textwidth]{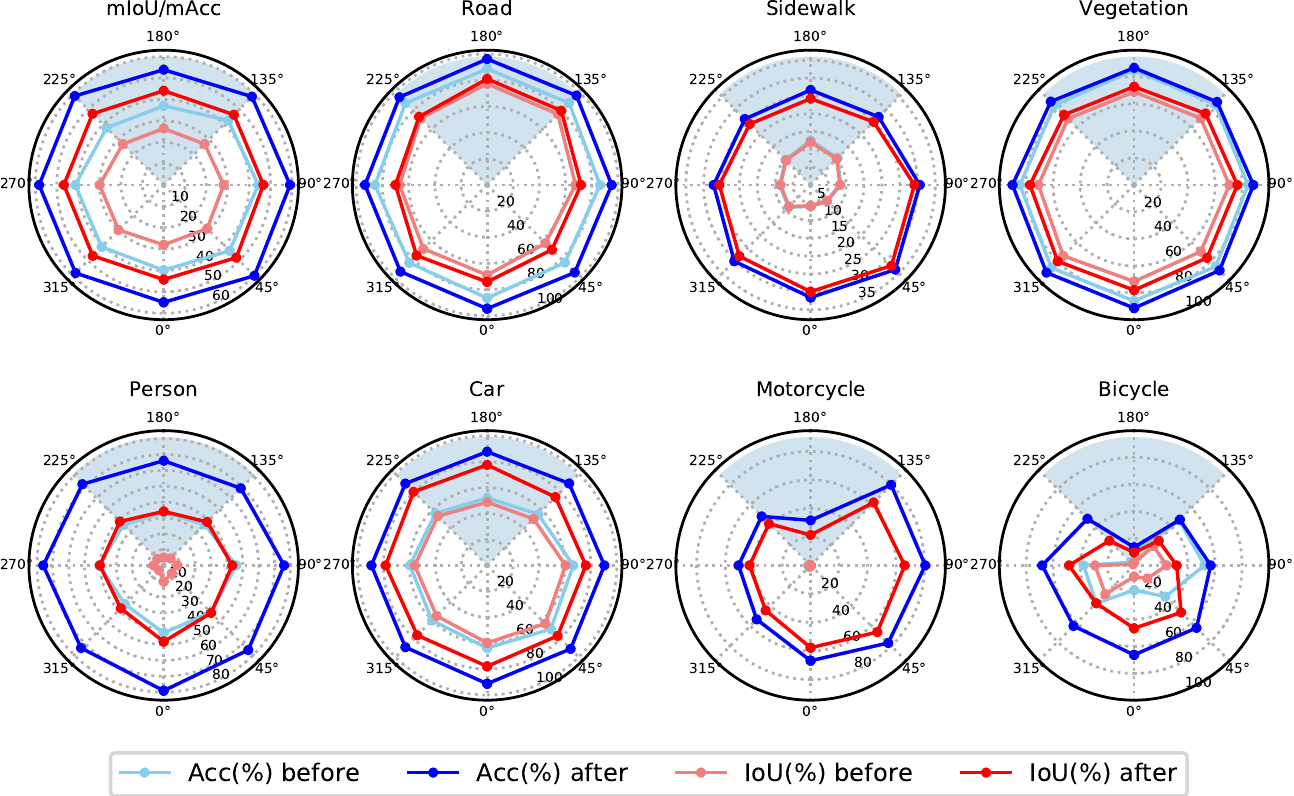}
  \vskip-1ex
  \caption{Class-wise Pixel Accuracy (Acc) and IoU comparison in different directions of the panoramic image, before and after adaptation.}
  \label{fig.polar_metrics}
  \vskip-3ex
\end{figure}

\begin{figure*}[!t]
  \centering
  \includegraphics[width= \textwidth]{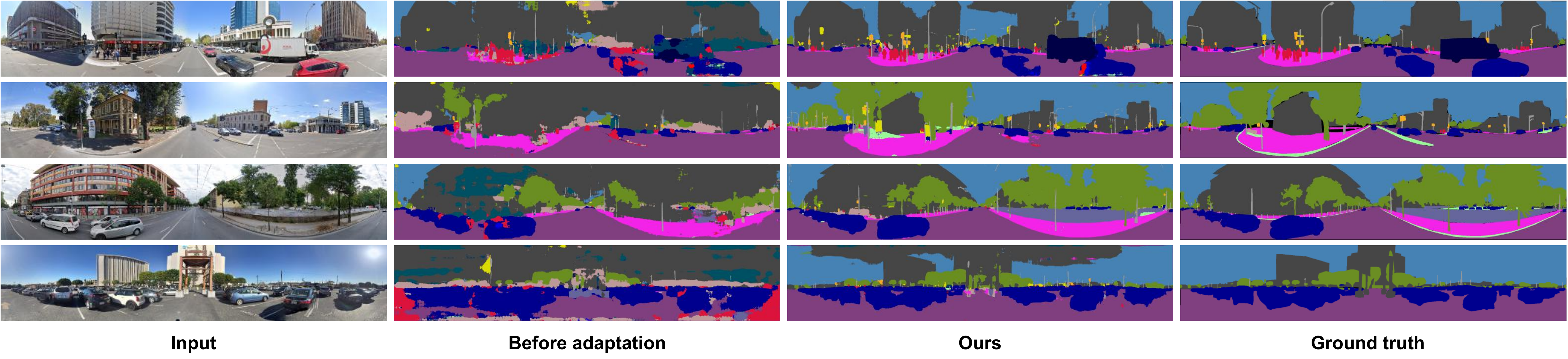}
  \vskip-1ex
  \caption{Qualitative examples of semantic segmentation on panoramic images. 
  The columns from left to right are the original input panoramic image, ERFNet predictions before adaptation, our predictions, and the ground truth. Zoom in for a better view.}
  \vskip-1ex
  \label{fig.comparision}
\end{figure*}

To broaden our comparison, we also consider several multi-supervision methods which benefit from multi-source data for a better generalization.
Seamless-Scene-Segmentation~\cite{seamless} uses instance segmentation labels for auxiliary supervision, whereas USSS~\cite{usss} performs multi-source semi-supervised learning.
The outputs of these models are mapped to the $19$ classes in DensePASS to be comparable with other models.
ISSAFE~\cite{issafe} merges multiple training datasets including Cityscapes, KITTI-360~\cite{kitti360}, and BDD~\cite{bdd} for safety-critical accident scene segmentation.
Our experiments indicate that  all these multi-source frameworks are sub-optimal in contrast to P2PDA which consistently leads to the best recognition rates. 
At the same time, P2PDA is trained on far less trainig data as the above approaches leverage larger databases, such as BDD/IDD and Mapillay, for training~\cite{issafe,omnirange}.
Especially for the classes \emph{building}, \emph{truck}, \emph{train}, and \emph{bicycle}, our framework is a front-runner by a large margin, as seen in Table~\ref{tab:more}.

To grasp the key prediction differences before and after the domain adaptation with P2PDA, we compare the Pixel Accuracy (Acc) and IoU in different directions of the panoramic image in Fig.~\ref{fig.polar_metrics}, where the blue-tinted regions indicate the section visible to a forward-facing narrow-FoV pinhole camera.
We partition the $360^\circ$ into $8$ directions and compute the class-wise accuracy of navigation-critical categories separately for each direction.
Our model leads to a considerable performance increase in all directions and for all the classes.
While the same panoramic view can be achieved from multiple cameras surrounding a vehicle, our system enables reliable deployment using a single camera together with good performances in certain safety-critical directions.
In particular, the recognition quality of \emph{sidewalk}, \emph{person}, and \emph{motorcycle} is improved by an especially large margin through the domain adaptation paradigm.
For the critical \emph{road} and \emph{car} segmentation relevant to autonomous driving, we have reached pixel accuracy at the level of $90\%$ around the $360^\circ$.

\subsection{Complementing the Cityscapes Source with WildDash}
\label{sec:wilddash}
Since most of the pinhole images are captured in the front view~\cite{cityscapes}, the features differ from the ones of $360^\circ$ panoramic images, which leads to the feature insufficiency of the side and rear views.
With the self-supervised learning phase, from high-confidence pseudo-labels, our model can learn more panoramic-oriented features including the side and rear views.
While the attentional- and regional adaptation modules already help address this issue by propagating features across the entire FoV and reduce the domain gap, to further complement the image feature from these perspectives, we consider exploiting a more diverse dataset in the P2PDA framework. 

Thereby, our next area of investigation is the impact of expanding the source data with a more complex dataset, since DensePASS contains highly composite scenes due to larger FoV, while Cityscapes is large but relatively simple.
To achieve this, we leverage the WildDash dataset~\cite{wilddash} with $4256$ pinhole images, pixel-level annotations and more unstructured surroundings.
For the training, we aggregate Cityscapes and WildDash sources without any complex joint training methodologies.
As shown in the last rows of Table~\ref{tab:danet}, we obtain better mIoU with the expanded training set, achieving $42.87\%$ and $44.66\%$ with different P2PDA variants.
Interestingly, the IoUs of \emph{road}, \emph{sidewalk}, \emph{fence}, \emph{terrain}, and \emph{car}
are significantly improved, which we link to the strong positional priors of these categories in structured urban scenes, while DensePASS and WildDash environment is more chaotic and unconstrained.
Moreover, direct comparison of the different P2PDA variants further verifies the effectiveness of the attention-augmented and uncertainty-aware adaptation strategies.

\begin{figure*}[!t]
  \centering
  \includegraphics[width= \textwidth]{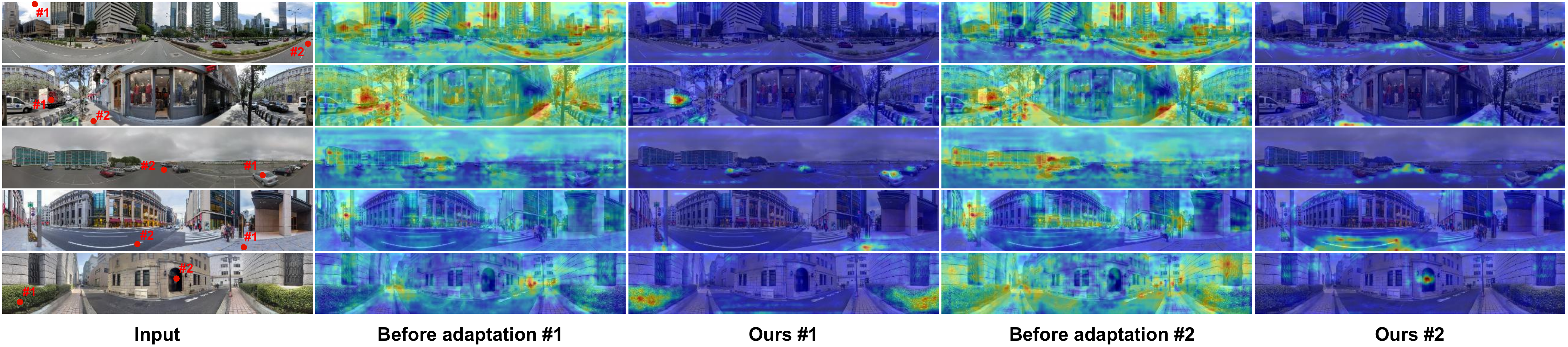}
  \vskip-1ex
  \caption{Visualization of attention maps from DANet before and after domain adaptation. For each input panorama, we select two points and show their corresponding position attention maps. Zoom in for a better view.}
  \vskip-1ex
  \label{fig.attention}
\end{figure*}

\begin{figure*}[!t]
  \centering
  \includegraphics[width= \textwidth]{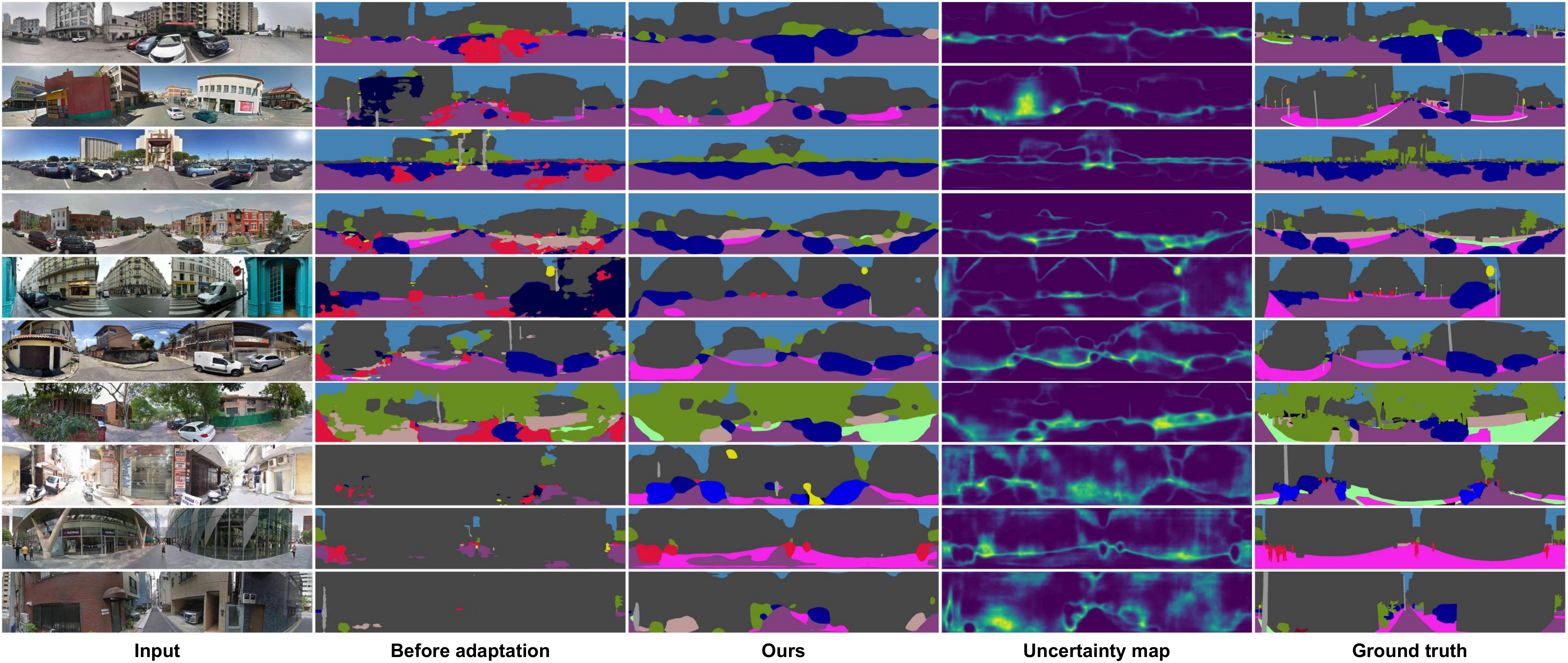}
  \vskip-1ex
  \caption{Qualitative examples of semantic segmentation on panoramic images. 
  From left to right columns are the input image, DANet predictions before adaptation, our predictions, the uncertainty map (brighter areas indicate higher uncertainty), and the ground truth. }
  \vskip-1ex
  \label{fig.comparision_danet}
\end{figure*}

\subsection{Qualitative Analysis}
\label{sec:qualitative}

In our final study, we showcase multiple examples of representative qualitative results in Fig.~\ref{fig.comparision}, Fig.~\ref{fig.attention}, and Fig.~\ref{fig.comparision_danet}.
Fig.~\ref{fig.comparision} displays the segmentation results of ERFNet.  
The segmentation boundaries of regions such as \emph{sky}, \emph{building}, and \emph{vegetation} are clearly improved through the P2PDA strategy in every case, while \emph{sidewalk} segmentation clearly benefits from domain adaptation in the second row example.
At the same time, some misclassified categories are corrected after adaptation (\textit{e.g.}, \emph{car} and \emph{truck} in all examples). 
There is also more clarity as it comes to detailed segmentation of small objects, such as \emph{traffic light} and \emph{traffic sign} in the first and the second row, as well as \emph{pole} in all rows. 
These qualitative examples consistently confirm the conclusions of our quantitative evaluation, highlighting the benefits of the proposed P2PDA strategy for $360^\circ$ self-driving scene understanding through attention-augmented adaptation from pinhole camera data. 

We now deepen the analysis and study the influence of our framework on the attention maps. As shown in Fig.~\ref{fig.attention}, for each input panorama, we select two points and visualize their corresponding position attention maps generated by DANet.
Prior to adaptation, the results are very chaotic and it is hard to find correct  meaningful correlations.
However, after adaptation, the network has learned to capture semantic associations with clear similarity and long-range dependencies that can stretch across the $360^\circ$.
For example, for the point on the \emph{car} in the third row, the attention map allows to focus and highlights other pixels of \emph{car}.
In the fourth row, the position attention successfully distinguishes \emph{road} and \emph{sidewalk}, which are crucial for self-driving applications. 
In the last row, for the point marked on \emph{terrain}, the corresponding attention map spotlights the other area of \emph{terrain} lying at a long distance.
In a nutshell, these  results further demonstrate that the attention-augmented P2PDA enables to seamlessly adapt to and effectively exploit semantic correlations in the panoramic imagery. 

We further demonstrate multiple predictions of our adapted DANet in Fig.~\ref{fig.comparision_danet}, showing a clear performance decline of the source-only DANet when applied on panoramic images.
In some of the top examples, the baseline often confuses the segmentation of some foreground categories, such as  \emph{cars}.
Even in the last three lines, it cannot distinguish other categories from the \emph{building} category in complex scenarios which is clearly better with the adapted DANet version.
Despite lacking sharp boundaries, the adapted model is superior at distinguishing the categories, which is particularly important for autonomous vehicles.
Strategies to augment the details include leveraging disentangled attention to handle detailed dependencies~\cite{omnirange} or directly using detail-sensitive networks~\cite{panopticfpn,dspass} for adaptation.
We want to mention, that while the uncertainty maps are mainly used to select high-confident pseudo-labels for the self-supervised learning, they could also be utilized as an attention cue for the assistance system during driving.

\section{Conclusion}

Semantic scene understanding is vital for self-driving cars but requires models which can deal with changes in data distribution.
Panoramic image segmentation is able to provide an entire $360^\circ$ surrounding perception. While appealing, it is challenging due to the discrepancy to common pinhole image segmentation in terms of the field of view and semantic distribution.
In  this  work,  we introduced the new task of cross-domain semantic segmentation for panoramic driving scenes, which extends the standard panoramic segmentation with the premise of the training data originating from a different domain (\textit{e.g.} pinhole camera images).

First, we formulate the problem of unsupervised domain adaptation for panoramic segmentation and introduce the novel DensePASS dataset which we use to study the \textsc{Pinhole$\rightarrow$Panoramic} transfer.
To meet the challenge of domain divergence, we developed a generic P2PDA framework enhancing conventional segmentation algorithms with different domain adaptation modules.
While our experiments demonstrate that cross-domain panoramic segmentation task is difficult for modern algorithms, our proposed domain-agnostic framework with attention-based and uncertainty-aware adaptation modules consistently improves the results.
Our dataset will be publicly released upon publication and we believe that DensePASS has strong  potential to motivate the needed development of \emph{generalizable} semantic segmentation models.

In the future, we aim to explore efficient transformer architectures and investigate their adaptation and distillation for semantic segmentation.
We intend to incorporate dense $360^\circ$ top-view LiDAR data panoramic segmentation to establish a more complete surrounding perception system and perform cross-view multi-dimension semantic mapping.
Further, we seek to expand the applicability of omnidirectional sensing and are particularly interested in aerial image segmentation by lifting panoramic scene parsing in drone videos.

\bibliographystyle{IEEEtran}
\bibliography{bib}

\begin{thebibliography}{10}
\providecommand{\url}[1]{#1}
\csname url@samestyle\endcsname
\providecommand{\newblock}{\relax}
\providecommand{\bibinfo}[2]{#2}
\providecommand{\BIBentrySTDinterwordspacing}{\spaceskip=0pt\relax}
\providecommand{\BIBentryALTinterwordstretchfactor}{4}
\providecommand{\BIBentryALTinterwordspacing}{\spaceskip=\fontdimen2\font plus
\BIBentryALTinterwordstretchfactor\fontdimen3\font minus
  \fontdimen4\font\relax}
\providecommand{\BIBforeignlanguage}[2]{{%
\expandafter\ifx\csname l@#1\endcsname\relax
\typeout{** WARNING: IEEEtran.bst: No hyphenation pattern has been}%
\typeout{** loaded for the language `#1'. Using the pattern for}%
\typeout{** the default language instead.}%
\else
\language=\csname l@#1\endcsname
\fi
#2}}
\providecommand{\BIBdecl}{\relax}
\BIBdecl

\bibitem{fcn}
J.~Long, E.~Shelhamer, and T.~Darrell, ``Fully convolutional networks for
  semantic segmentation,'' in \emph{2015 IEEE Conference on Computer Vision and
  Pattern Recognition (CVPR)}, 2015, pp. 3431--3440.

\bibitem{peng2021mass}
K.~Peng \emph{et~al.}, ``{MASS:} {Multi-attentional} semantic segmentation of
  {LiDAR} data for dense top-view understanding,'' \emph{arXiv preprint
  arXiv:2107.00346}, 2021.

\bibitem{cityscapes}
M.~Cordts \emph{et~al.}, ``The cityscapes dataset for semantic urban scene
  understanding,'' in \emph{2016 IEEE Conference on Computer Vision and Pattern
  Recognition (CVPR)}, 2016, pp. 3213--3223.

\bibitem{restricted}
L.~Deng, M.~Yang, H.~Li, T.~Li, B.~Hu, and C.~Wang, ``Restricted deformable
  convolution-based road scene semantic segmentation using surround view
  cameras,'' \emph{IEEE Transactions on Intelligent Transportation Systems},
  vol.~21, no.~10, pp. 4350--4362, 2020.

\bibitem{camera_lidar}
J.~S. Berrio, M.~Shan, S.~Worrall, and E.~Nebot, ``{Camera-LIDAR} integration:
  Probabilistic sensor fusion for semantic mapping,'' \emph{IEEE Transactions
  on Intelligent Transportation Systems}, 2021.

\bibitem{can_we_pass}
K.~Yang \emph{et~al.}, ``Can we {PASS} beyond the field of view? {P}anoramic
  annular semantic segmentation for real-world surrounding perception,'' in
  \emph{2019 IEEE Intelligent Vehicles Symposium (IV)}, 2019, pp. 446--453.

\bibitem{pass}
K.~Yang, X.~Hu, L.~M. Bergasa, E.~Romera, and K.~Wang, ``{PASS:} {P}anoramic
  annular semantic segmentation,'' \emph{IEEE Transactions on Intelligent
  Transportation Systems}, vol.~21, no.~10, pp. 4171--4185, 2020.

\bibitem{bridging}
E.~Romera, L.~M. Bergasa, K.~Yang, J.~M. Alvarez, and R.~Barea, ``Bridging the
  day and night domain gap for semantic segmentation,'' in \emph{2019 IEEE
  Intelligent Vehicles Symposium (IV)}, 2019, pp. 1312--1318.

\bibitem{issafe}
J.~Zhang, K.~Yang, and R.~Stiefelhagen, ``{ISSAFE:} {I}mproving semantic
  segmentation in accidents by fusing event-based data,'' in \emph{2021
  IEEE/RSJ International Conference on Intelligent Robots and Systems (IROS)},
  2021, pp. 1--8.

\bibitem{adaptsegnet}
Y.-H. Tsai, W.-C. Hung, S.~Schulter, K.~Sohn, M.-H. Yang, and M.~Chandraker,
  ``Learning to adapt structured output space for semantic segmentation,'' in
  \emph{2018 IEEE/CVF Conference on Computer Vision and Pattern Recognition
  (CVPR)}, 2018, pp. 7472--7481.

\bibitem{all_about_structure}
W.-L. Chang, H.-P. Wang, W.-H. Peng, and W.-C. Chiu, ``All about structure:
  Adapting structural information across domains for boosting semantic
  segmentation,'' in \emph{2019 IEEE/CVF Conference on Computer Vision and
  Pattern Recognition (CVPR)}, 2019, pp. 1900--1909.

\bibitem{nonlocal}
X.~Wang, R.~Girshick, A.~Gupta, and K.~He, ``Non-local neural networks,'' in
  \emph{2018 IEEE/CVF Conference on Computer Vision and Pattern Recognition
  (CVPR)}, 2018, pp. 7794--7803.

\bibitem{danet}
J.~Fu \emph{et~al.}, ``Dual attention network for scene segmentation,'' in
  \emph{2019 IEEE/CVF Conference on Computer Vision and Pattern Recognition
  (CVPR)}, 2019, pp. 3141--3149.

\bibitem{fanet}
P.~Hu \emph{et~al.}, ``Real-time semantic segmentation with fast attention,''
  \emph{IEEE Robotics and Automation Letters}, vol.~6, no.~1, pp. 263--270,
  2021.

\bibitem{erfnet}
E.~Romera, J.~M. Alvarez, L.~M. Bergasa, and R.~Arroyo, ``{ERFNet:} {E}fficient
  residual factorized {ConvNet} for real-time semantic segmentation,''
  \emph{IEEE Transactions on Intelligent Transportation Systems}, vol.~19,
  no.~1, pp. 263--272, 2018.

\bibitem{densepass_itsc}
C.~Ma, J.~Zhang, K.~Yang, A.~Roitberg, and R.~Stiefelhagen, ``{DensePASS:}
  {Dense} panoramic semantic segmentation via unsupervised domain adaptation
  with attention-augmented context exchange,'' in \emph{2021 IEEE 24th
  International Conference on Intelligent Transportation Systems (ITSC)}, 2021,
  pp. 1--7.

\bibitem{wilddash}
O.~Zendel, K.~Honauer, M.~Murschitz, D.~Steininger, and G.~F. Dom{\'\i}nguez,
  ``{WildDash} - {C}reating hazard-aware benchmarks,'' in \emph{European
  Conference on Computer Vision}, 2018, pp. 407--421.

\bibitem{deep_multimodal}
D.~Feng \emph{et~al.}, ``Deep multi-modal object detection and semantic
  segmentation for autonomous driving: Datasets, methods, and challenges,''
  \emph{IEEE Transactions on Intelligent Transportation Systems}, vol.~22,
  no.~3, pp. 1341--1360, 2021.

\bibitem{pspnet}
H.~Zhao, J.~Shi, X.~Qi, X.~Wang, and J.~Jia, ``Pyramid scene parsing network,''
  in \emph{2017 IEEE Conference on Computer Vision and Pattern Recognition
  (CVPR)}, 2017, pp. 6230--6239.

\bibitem{deeplabv3+}
L.-C. Chen, Y.~Zhu, G.~Papandreou, F.~Schroff, and H.~Adam, ``Encoder-decoder
  with atrous separable convolution for semantic image segmentation,'' in
  \emph{European Conference on Computer Vision}, 2018, pp. 833--851.

\bibitem{panopticfpn}
A.~Kirillov, R.~Girshick, K.~He, and P.~Doll{\'a}r, ``Panoptic feature pyramid
  networks,'' in \emph{2019 IEEE/CVF Conference on Computer Vision and Pattern
  Recognition (CVPR)}, 2019, pp. 6392--6401.

\bibitem{swiftnet}
M.~Orsic, I.~Kreso, P.~Bevandic, and S.~Segvic, ``In defense of pre-trained
  imagenet architectures for real-time semantic segmentation of road-driving
  images,'' in \emph{2019 IEEE/CVF Conference on Computer Vision and Pattern
  Recognition (CVPR)}, 2019, pp. 12\,599--12\,608.

\bibitem{fastscnn}
R.~P.~K. Poudel, S.~Liwicki, and R.~Cipolla, ``{Fast-SCNN:} {F}ast semantic
  segmentation network,'' in \emph{British Machine Vision Conference}, 2019, p.
  289.

\bibitem{mapillary}
G.~Neuhold, T.~Ollmann, S.~R. Bul{\`o}, and P.~Kontschieder, ``The mapillary
  vistas dataset for semantic understanding of street scenes,'' in \emph{2017
  IEEE International Conference on Computer Vision (ICCV)}, 2017, pp.
  5000--5009.

\bibitem{apolloscape}
X.~Huang, P.~Wang, X.~Cheng, D.~Zhou, Q.~Geng, and R.~Yang, ``The {ApolloScape}
  open dataset for autonomous driving and its application,'' \emph{IEEE
  Transactions on Pattern Analysis and Machine Intelligence}, vol.~42, no.~10,
  pp. 2702--2719, 2020.

\bibitem{bdd}
F.~Yu \emph{et~al.}, ``{BDD100K:} {A} diverse driving dataset for heterogeneous
  multitask learning,'' in \emph{2020 IEEE/CVF Conference on Computer Vision
  and Pattern Recognition (CVPR)}, 2020, pp. 2633--2642.

\bibitem{idd}
G.~Varma, A.~Subramanian, A.~Namboodiri, M.~Chandraker, and C.~Jawahar,
  ``{IDD:} {A} dataset for exploring problems of autonomous navigation in
  unconstrained environments,'' in \emph{2019 IEEE Winter Conference on
  Applications of Computer Vision (WACV)}, 2019, pp. 1743--1751.

\bibitem{kitti360}
J.~Xie, M.~Kiefel, M.-T. Sun, and A.~Geiger, ``Semantic instance annotation of
  street scenes by {3D} to {2D} label transfer,'' in \emph{2016 IEEE Conference
  on Computer Vision and Pattern Recognition (CVPR)}, 2016, pp. 3688--3697.

\bibitem{attention}
A.~Vaswani \emph{et~al.}, ``Attention is all you need,'' \emph{Advances in
  Neural Information Processing Systems}, vol.~30, pp. 5998--6008, 2017.

\bibitem{zhang2021trans4trans}
J.~Zhang, K.~Yang, A.~Constantinescu, K.~Peng, K.~M{\"u}ller, and
  R.~Stiefelhagen, ``{Trans4Trans:} {Efficient} transformer for transparent
  object segmentation to help visually impaired people navigate in the real
  world,'' in \emph{2021 IEEE/CVF International Conference on Computer Vision
  Workshops (ICCVW)}, 2021, pp. 1760--1770.

\bibitem{ranet}
D.~Shen, Y.~Ji, P.~Li, Y.~Wang, and D.~Lin, ``{RANet:} {R}egion attention
  network for semantic segmentation,'' \emph{Advances in Neural Information
  Processing Systems}, vol.~33, pp. 1--12, 2020.

\bibitem{ocrnet}
Y.~Yuan, X.~Chen, and J.~Wang, ``Object-contextual representations for semantic
  segmentation,'' in \emph{European Conference on Computer Vision}, 2020, pp.
  173--190.

\bibitem{omnirange}
K.~Yang, J.~Zhang, S.~Rei{\ss}, X.~Hu, and R.~Stiefelhagen, ``Capturing
  omni-range context for omnidirectional segmentation,'' in \emph{2021 IEEE/CVF
  Conference on Computer Vision and Pattern Recognition (CVPR)}, 2021, pp.
  1376--1386.

\bibitem{attanet}
Q.~Song, K.~Mei, and R.~Huang, ``{AttaNet:} {A}ttention-augmented network for
  fast and accurate scene parsing,'' in \emph{Conference on Artificial
  Intelligence}, 2021, pp. 2567--2575.

\bibitem{ccnet}
Z.~Huang, X.~Wang, L.~Huang, C.~Huang, Y.~Wei, and W.~Liu, ``{CCNet:}
  {C}riss-cross attention for semantic segmentation,'' in \emph{2019 IEEE/CVF
  International Conference on Computer Vision (ICCV)}, 2019, pp. 603--612.

\bibitem{dnl}
M.~Yin \emph{et~al.}, ``Disentangled non-local neural networks,'' in
  \emph{European Conference on Computer Vision}, 2020, pp. 191--207.

\bibitem{annn}
Z.~Zhu, M.~Xu, S.~Bai, T.~Huang, and X.~Bai, ``Asymmetric non-local neural
  networks for semantic segmentation,'' in \emph{2019 IEEE/CVF International
  Conference on Computer Vision (ICCV)}, 2019, pp. 593--602.

\bibitem{fisheye}
L.~Deng, M.~Yang, Y.~Qian, C.~Wang, and B.~Wang, ``{CNN} based semantic
  segmentation for urban traffic scenes using fisheye camera,'' in \emph{2017
  IEEE Intelligent Vehicles Symposium (IV)}, 2017, pp. 231--236.

\bibitem{universal}
Y.~Ye, K.~Yang, K.~Xiang, J.~Wang, and K.~Wang, ``Universal semantic
  segmentation for fisheye urban driving images,'' in \emph{2020 IEEE
  International Conference on Systems, Man, and Cybernetics (SMC)}, 2020, pp.
  648--655.

\bibitem{adaptable_deformable}
C.~Playout, O.~Ahmad, F.~Lecue, and F.~Cheriet, ``Adaptable deformable
  convolutions for semantic segmentation of fisheye images in autonomous
  driving systems,'' \emph{arXiv preprint arXiv:2102.10191}, 2021.

\bibitem{crossview}
B.~Pan, J.~Sun, H.~Y.~T. Leung, A.~Andonian, and B.~Zhou, ``Cross-view semantic
  segmentation for sensing surroundings,'' \emph{IEEE Robotics and Automation
  Letters}, vol.~5, no.~3, pp. 4867--4873, 2020.

\bibitem{omnidet}
V.~Ravikumar \emph{et~al.}, ``{OmniDet:} {S}urround view cameras based
  multi-task visual perception network for autonomous driving,'' \emph{IEEE
  Robotics and Automation Letters}, vol.~6, no.~2, pp. 2830--2837, 2021.

\bibitem{hu2021fiery}
A.~Hu \emph{et~al.}, ``{FIERY:} {Future} instance prediction in bird's-eye view
  from surround monocular cameras,'' in \emph{2021 IEEE/CVF International
  Conference on Computer Vision (ICCV)}, 2021, pp. 15\,273--15\,282.

\bibitem{dspass}
K.~Yang, X.~Hu, H.~Chen, K.~Xiang, K.~Wang, and R.~Stiefelhagen, ``{DS-PASS:}
  {D}etail-sensitive panoramic annular semantic segmentation through {SwaftNet}
  for surrounding sensing,'' in \emph{2020 IEEE Intelligent Vehicles Symposium
  (IV)}, 2020, pp. 457--464.

\bibitem{ooss}
K.~Yang, X.~Hu, Y.~Fang, K.~Wang, and R.~Stiefelhagen, ``Omnisupervised
  omnidirectional semantic segmentation,'' \emph{IEEE Transactions on
  Intelligent Transportation Systems}, 2020.

\bibitem{wildpass}
K.~Yang, X.~Hu, and R.~Stiefelhagen, ``Is context-aware {CNN} ready for the
  surroundings? {P}anoramic semantic segmentation in the wild,'' \emph{IEEE
  Transactions on Image Processing}, vol.~30, pp. 1866--1881, 2021.

\bibitem{pps}
A.~Jaus, K.~Yang, and R.~Stiefelhagen, ``Panoramic panoptic segmentation:
  Towards complete surrounding understanding via unsupervised contrastive
  learning,'' in \emph{2021 IEEE Intelligent Vehicles Symposium (IV)}, 2021,
  pp. 1--7.

\bibitem{orientation}
C.~Zhang, S.~Liwicki, W.~Smith, and R.~Cipolla, ``Orientation-aware semantic
  segmentation on icosahedron spheres,'' in \emph{2019 IEEE/CVF International
  Conference on Computer Vision (ICCV)}, 2019, pp. 3532--3540.

\bibitem{synthetic}
Y.~Xu, K.~Wang, K.~Yang, D.~Sun, and J.~Fu, ``Semantic segmentation of
  panoramic images using a synthetic dataset,'' in \emph{Artificial
  Intelligence and Machine Learning in Defense Applications}, 2019.

\bibitem{omniscape}
A.~R. Sekkat, Y.~Dupuis, P.~Vasseur, and P.~Honeine, ``The {OmniScape}
  dataset,'' in \emph{2020 IEEE International Conference on Robotics and
  Automation (ICRA)}, 2020, pp. 1603--1608.

\bibitem{cortinhal2021semantics}
T.~Cortinhal, F.~Kurnaz, and E.~Aksoy, ``Semantics-aware multi-modal domain
  translation: From {LiDAR} point clouds to panoramic color images,''
  \emph{arXiv preprint arXiv:2106.13974}, 2021.

\bibitem{orhan2021semantic}
S.~Orhan and Y.~Bastanlar, ``Semantic segmentation of outdoor panoramic
  images,'' \emph{Signal, Image and Video Processing}, pp. 1--8, 2021.

\bibitem{see_clearer_at_night}
L.~Sun, K.~Wang, K.~Yang, and K.~Xiang, ``See clearer at night: {T}owards
  robust nighttime semantic segmentation through day-night image conversion,''
  in \emph{Artificial Intelligence and Machine Learning in Defense
  Applications}, 2019.

\bibitem{nighttime_road_scene_parsing}
C.~Song, J.~Wu, L.~Zhu, M.~Zhang, and H.~Ling, ``Nighttime road scene parsing
  by unsupervised domain adaptation,'' \emph{IEEE Transactions on Intelligent
  Transportation Systems}, 2020.

\bibitem{rainy_night}
S.~Di \emph{et~al.}, ``Rainy night scene understanding with near scene semantic
  adaptation,'' \emph{IEEE Transactions on Intelligent Transportation Systems},
  vol.~22, no.~3, pp. 1594--1602, 2021.

\bibitem{curriculum_da}
Y.~Zhang, P.~David, and B.~Gong, ``Curriculum domain adaptation for semantic
  segmentation of urban scenes,'' in \emph{2017 IEEE International Conference
  on Computer Vision (ICCV)}, 2017, pp. 2039--2049.

\bibitem{pycda}
Q.~Lian, L.~Duan, F.~Lv, and B.~Gong, ``Constructing self-motivated pyramid
  curriculums for cross-domain semantic segmentation: A non-adversarial
  approach,'' in \emph{2019 IEEE/CVF International Conference on Computer
  Vision (ICCV)}, 2019, pp. 6757--6766.

\bibitem{crst}
Y.~Zou, Z.~Yu, X.~Liu, B.~V. Kumar, and J.~Wang, ``Confidence regularized
  self-training,'' in \emph{2019 IEEE/CVF International Conference on Computer
  Vision (ICCV)}, 2019, pp. 5981--5990.

\bibitem{rectifying}
Z.~Zheng and Y.~Yang, ``Rectifying pseudo label learning via uncertainty
  estimation for domain adaptive semantic segmentation,'' \emph{International
  Journal of Computer Vision}, vol. 129, pp. 1106--1120, 2021.

\bibitem{proda}
P.~Zhang, B.~Zhang, T.~Zhang, D.~Chen, Y.~Wang, and F.~Wen, ``Prototypical
  pseudo label denoising and target structure learning for domain adaptive
  semantic segmentation,'' in \emph{2021 IEEE/CVF Conference on Computer Vision
  and Pattern Recognition (CVPR)}, 2021, pp. 12\,414--12\,424.

\bibitem{fcnsinthewild}
J.~Hoffman, D.~Wang, F.~Yu, and T.~Darrell, ``{FCNs} in the wild: {P}ixel-level
  adversarial and constraint-based adaptation,'' \emph{arXiv preprint
  arXiv:1612.02649}, 2016.

\bibitem{cycada}
J.~Hoffman \emph{et~al.}, ``{CyCADA:} {C}ycle-consistent adversarial domain
  adaptation,'' in \emph{International Conference on Machine Learning}, 2018,
  pp. 1989--1998.

\bibitem{clan}
Y.~Luo, L.~Zheng, T.~Guan, J.~Yu, and Y.~Yang, ``Taking a closer look at domain
  shift: Category-level adversaries for semantics consistent domain
  adaptation,'' in \emph{2019 IEEE/CVF Conference on Computer Vision and
  Pattern Recognition (CVPR)}, 2019, pp. 2502--2511.

\bibitem{gan}
I.~Goodfellow \emph{et~al.}, ``Generative adversarial nets,'' \emph{Advances in
  Neural Information Processing Systems}, vol.~27, pp. 2672--2680, 2014.

\bibitem{content_consistent_matching_da}
G.~Li, G.~Kang, W.~Liu, Y.~Wei, and Y.~Yang, ``Content-consistent matching for
  domain adaptive semantic segmentation,'' in \emph{European Conference on
  Computer Vision}, 2020, pp. 440--456.

\bibitem{contextual_relation_consistent_da}
J.~Huang, S.~Lu, D.~Guan, and X.~Zhang, ``Contextual-relation consistent domain
  adaptation for semantic segmentation,'' in \emph{European Conference on
  Computer Vision}, 2020, pp. 705--722.

\bibitem{advent}
T.-H. Vu, H.~Jain, M.~Bucher, M.~Cord, and P.~P{\'e}rez, ``{ADVENT:}
  {A}dversarial entropy minimization for domain adaptation in semantic
  segmentation,'' in \emph{2019 IEEE/CVF Conference on Computer Vision and
  Pattern Recognition (CVPR)}, 2019, pp. 2512--2521.

\bibitem{affinity_space_da}
W.~Zhou, Y.~Wang, J.~Chu, J.~Yang, X.~Bai, and Y.~Xu, ``Affinity space
  adaptation for semantic segmentation across domains,'' \emph{IEEE
  Transactions on Image Processing}, vol.~30, pp. 2549--2561, 2021.

\bibitem{tada}
X.~Wang, L.~Li, W.~Ye, M.~Long, and J.~Wang, ``Transferable attention for
  domain adaptation,'' in \emph{Conference on Artificial Intelligence}, 2019,
  pp. 5345--5352.

\bibitem{safe}
D.~Kothandaraman, R.~Chandra, and D.~Manocha, ``{SS-SFDA} : {S}elf-supervised
  source-free domain adaptation for road segmentation in hazardous
  environments,'' \emph{arXiv preprint arXiv:2012.08939}, 2020.

\bibitem{source_free}
Y.~Liu, W.~Zhang, and J.~Wang, ``Source-free domain adaptation for semantic
  segmentation,'' in \emph{2021 IEEE/CVF Conference on Computer Vision and
  Pattern Recognition (CVPR)}, 2021, pp. 1215--1224.

\bibitem{contextaware}
J.~Yang, W.~An, C.~Yan, P.~Zhao, and J.~Huang, ``Context-aware domain
  adaptation in semantic segmentation,'' in \emph{2021 IEEE/CVF Winter
  Conference on Applications of Computer Vision (WACV)}, 2021, pp. 514--524.

\bibitem{gong2020mdalu}
R.~Gong, D.~Dai, Y.~Chen, W.~Li, and L.~Van~Gool, ``{mDALU:} {Multi-source}
  domain adaptation and label unification with partial datasets,'' in
  \emph{2021 IEEE/CVF International Conference on Computer Vision (ICCV)},
  2021, pp. 8876--8885.

\bibitem{adam}
D.~P. Kingma and J.~Ba, ``Adam: A method for stochastic optimization,'' in
  \emph{International Conference on Learning Representations}, 2015.

\bibitem{resnest}
H.~Zhang \emph{et~al.}, ``{ResNeSt:} {S}plit-attention networks,'' \emph{arXiv
  preprint arXiv:2004.08955}, 2020.

\bibitem{setr}
S.~Zheng \emph{et~al.}, ``Rethinking semantic segmentation from a
  sequence-to-sequence perspective with transformers,'' in \emph{2021 IEEE/CVF
  Conference on Computer Vision and Pattern Recognition (CVPR)}, 2021, pp.
  6881--6890.

\bibitem{seamless}
L.~Porzi, S.~R. Bul{\`o}, A.~Colovic, and P.~Kontschieder, ``Seamless scene
  segmentation,'' in \emph{2019 IEEE/CVF Conference on Computer Vision and
  Pattern Recognition (CVPR)}, 2019, pp. 8269--8278.

\bibitem{usss}
T.~Kalluri, G.~Varma, M.~Chandraker, and C.~Jawahar, ``Universal
  semi-supervised semantic segmentation,'' in \emph{2019 IEEE/CVF International
  Conference on Computer Vision (ICCV)}, 2019, pp. 5258--5269.

\end{thebibliography}

\end{document}